\documentclass[preprint,12pt]{elsarticle}

\usepackage{amssymb}
\usepackage{amsmath}
\usepackage[table,xcdraw]{xcolor}
\usepackage{tabularx}
\usepackage{adjustbox}
\usepackage{algorithm}
\usepackage{algorithmic}
\usepackage{changepage}
\usepackage{soul}       
\usepackage{siunitx}
\usepackage{multirow}
\usepackage{graphicx}
\usepackage{xcolor}
\usepackage{overpic}

\journal{Nuclear Physics B}

\begin{document}

\begin{frontmatter}



\title{Volume-Consistent Kneading-Based Deformation Manufacturing for Material-Efficient Shaping}


\author[aff1]{Li~Lei\corref{equal}}
\author[aff2]{Gong~Jiale\corref{equal}}
\author[aff1]{Li~Ziyang}
\author[aff1]{Wang~Hong\corref{cor}}

\affiliation[aff1]{organization={School of Mechanical Engineering and Automation, Northeastern University},
            city={Shenyang},
            postcode={110819},
            country={China}}

\affiliation[aff2]{organization={Future Laboratory, Tsinghua University},
            city={Beijing},
            postcode={100084},
            country={China}}

\cortext[equal]{These authors contributed equally to this work.}
\cortext[cor]{Corresponding author.}

\ead[cor]{hongwang@mail.neu.edu.cn}

\begin{abstract}
Conventional subtractive manufacturing inevitably involves material loss during geometric realization, while additive manufacturing still suffers from limitations in surface quality, process continuity, and productivity when fabricating complex geometries. To address these challenges, this paper proposes a volume-consistent kneading-based forming method for plastic materials, enabling continuous and controllable three-dimensional deformation under mass conservation. An integrated kneading-based manufacturing system is developed, in which geometry-aware kneading command generation, layer-wise kneading execution, and in-process point-cloud scanning are tightly coupled to form a closed-loop workflow of scanning, forming, and feedback compensation. Target geometries are analyzed through layer-wise point-cloud processing and classified into enveloping and non-enveloping types. Accordingly, an Envelope Shaping First strategy and a Similar Gradient Method are adopted to ensure stable material flow and continuous deformation. An RMSE-based compensation scheme is further introduced to correct systematic geometric deviations induced by elastic rebound and material redistribution. Experimental validation on five representative geometries demonstrates high geometric fidelity, with material utilization consistently exceeding 98\%. The results indicate that kneading-based forming provides a promising alternative manufacturing paradigm for low-waste, customizable production.
\end{abstract}



\end{frontmatter}

\section{Introduction}

Manufacturing systems have evolved from craft-oriented production to highly systematized and customization-oriented paradigms~\cite{Demand-integrated-flexible-MS}, where digitalization and flexibility increasingly reshape how manufacturing tasks are organized and optimized~\cite{mourtzis2014evolution}. Within this landscape, subtractive manufacturing remains a mature and widely deployed mechanism for geometric realization through controlled material removal, and has further expanded toward robotic machining and the processing of advanced materials such as composites~\cite{Sub-manu-robtic}. However, subtractive processes inherently involve material loss and potential environmental burdens; in the context of sustainable manufacturing, the trade-off between high precision and high waste becomes increasingly critical~\cite{Subtractive-versus-mass,Subtractive-Hazardous-Materials}.

Additive manufacturing significantly enlarges the feasible design space for complex components, and emerging directions such as multi-axis additive manufacturing~\cite{Cooperative-robotics-AM} and additive manufacturing of composite materials continue to broaden its applicability~\cite{Additive-manufactur-review,Review-multi-axis-AM}. Nevertheless, additive processes still face practical bottlenecks, including surface quality limitations, support-structure dependency, and distortion control challenges, which collectively increase process-chain complexity and constrain productivity and energy efficiency~\cite{Surface-roughness-AM,Support-structures-3D,Review-assessment-AM,Fabrication-opt-min-distortion}. As a result, despite the progress of both paradigms, there remains a persistent need for a manufacturing mechanism that can achieve continuous, controllable deformation while maintaining high material efficiency.

Under the Industry~4.0 framework, interconnectivity, data, and automation have enhanced equipment capability and system integration~\cite{Digital-twin-4.0}; yet, for personalized and complex products, manufacturing systems increasingly demand ``controllable deformation with real-time correction'' to maintain digital--physical consistency throughout production~\cite{Ten-Industrie-4.0,Architectural-model-4.0,Product-complexity-operational}. Digital twin technologies have been widely adopted to bridge design--process--quality information and to support knowledge modeling and in-process quality closure at the system level~\cite{Digital-twin-AM,sustainable-mass-customization,pattern-mining-supporting}. Meanwhile, error formation in equipment and processes often follows deterministic and thermal-driven mechanisms, making compensation strategies and data-driven models essential for repeatability and accuracy in integrated manufacturing workflows~\cite{Digital-robotic-machining,Deterministic-error-accuracy,Digital-Thermal-Error}. In addition, manufacturing data standards and process planning still face alignment challenges across the ``model--process--equipment'' chain, which limits traceability and automation across stages~\cite{STEP-NC-In-AM,Pro-plan-AM-SM,Review-sub-add-manufacturing}.

\begin{figure}[h]
\centering
\includegraphics[width=\columnwidth]{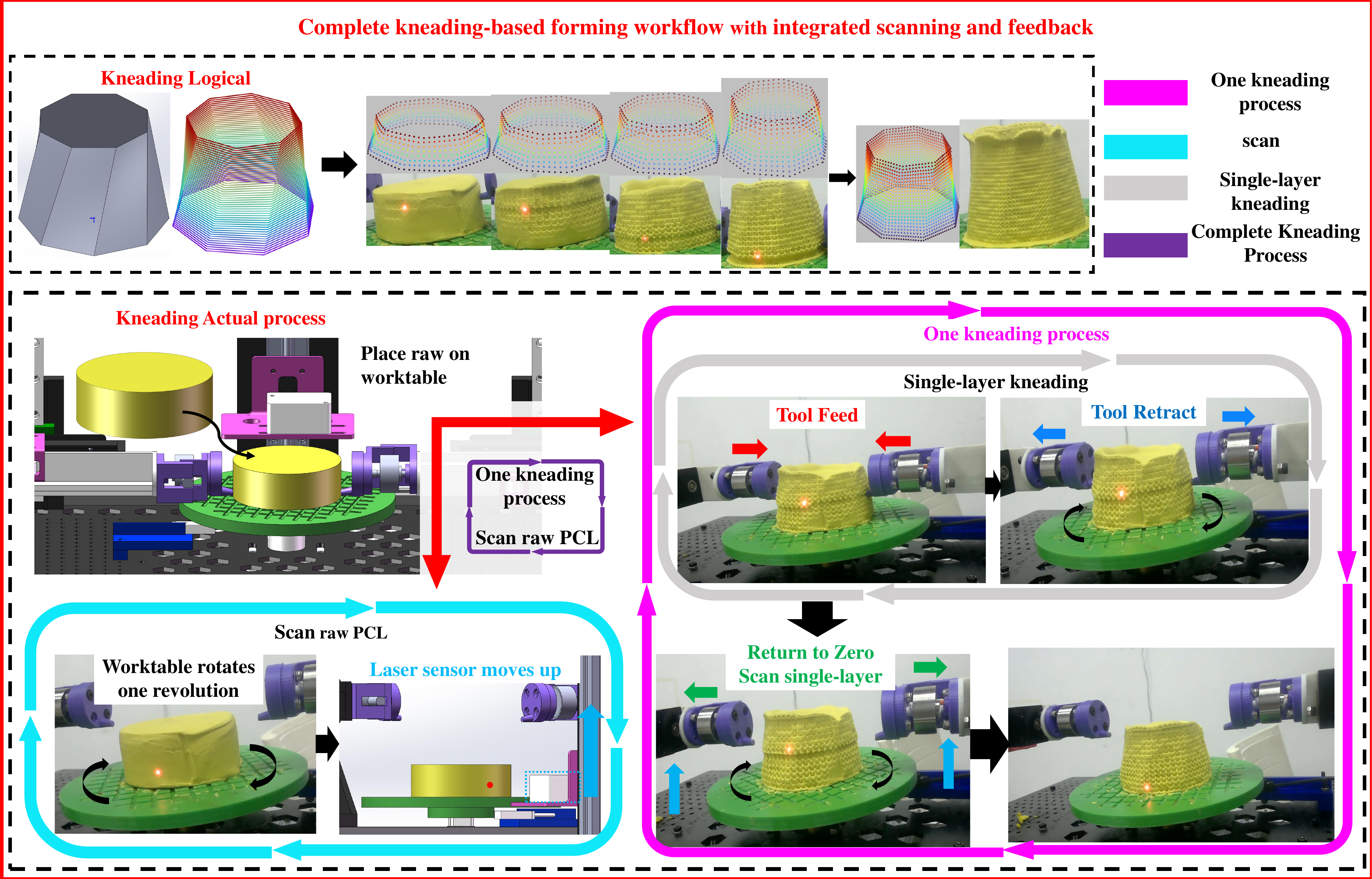}
\caption{Complete kneading-based forming workflow with integrated scanning and feedback.}
\label{complete_kneading_process}
\end{figure}

From the perspective of zero-defect and zero-waste manufacturing, non-destructive inspection and online measurement~\cite{3D-digital-deformation} provide the necessary basis for a closed-loop workflow of ``detecting deviation--feedback compensation--re-manufacturing''~\cite{Towards-Zero-Waste,Studies-Metrological-3D-scan}. Distinct from material removal or deposition, plastic forming emphasizes controlled material flow under volume (mass) conservation, where flow laws and control directly determine the attainable geometric accuracy~\cite{Advances-forming-metals,Advancements-extrusion-drawing,overview-continuous-severe-deformation,Material-Flow-Control}. In parallel, recent studies on robotic manipulation of deformable objects suggest that stability and controllability of shaping soft/plastic materials can be achieved through structural priors and planning strategies, providing inspiration for a computable and feedback-enabled volume-conserving shaping mechanism~\cite{Ropotter-deformable-manipulation,Planning-3D-Deformable}. To address the above challenges, this study proposes a kneading-based forming paradigm that enables continuous and controllable deformation of plastic materials under volume conservation. As illustrated in Fig.~\ref{complete_kneading_process}, the proposed manufacturing workflow integrates geometry-aware planning, layer-wise kneading execution, and in-process scanning into a closed-loop forming system. Raw material geometry is first captured through non-destructive scanning, followed by iterative single-layer kneading and re-scanning to compensate for geometric deviations induced by material flow and elastic rebound. By tightly coupling digital geometric representation with physical forming actions, the workflow ensures digital–physical consistency while achieving high material efficiency. 

\section{System and 3D model design}

Independently designed and built the mechanical structure, sensor system, and control system of the intelligent kneading device; developed the lower computer kneading controller software system, the upper computer software system for point cloud sampling, processing, and comparative analysis, as well as the automatic kneading instruction generation software system.

\subsection{Hardware system and software system design}

\begin{figure}[H]
\centering
\includegraphics[width=1.0\columnwidth]{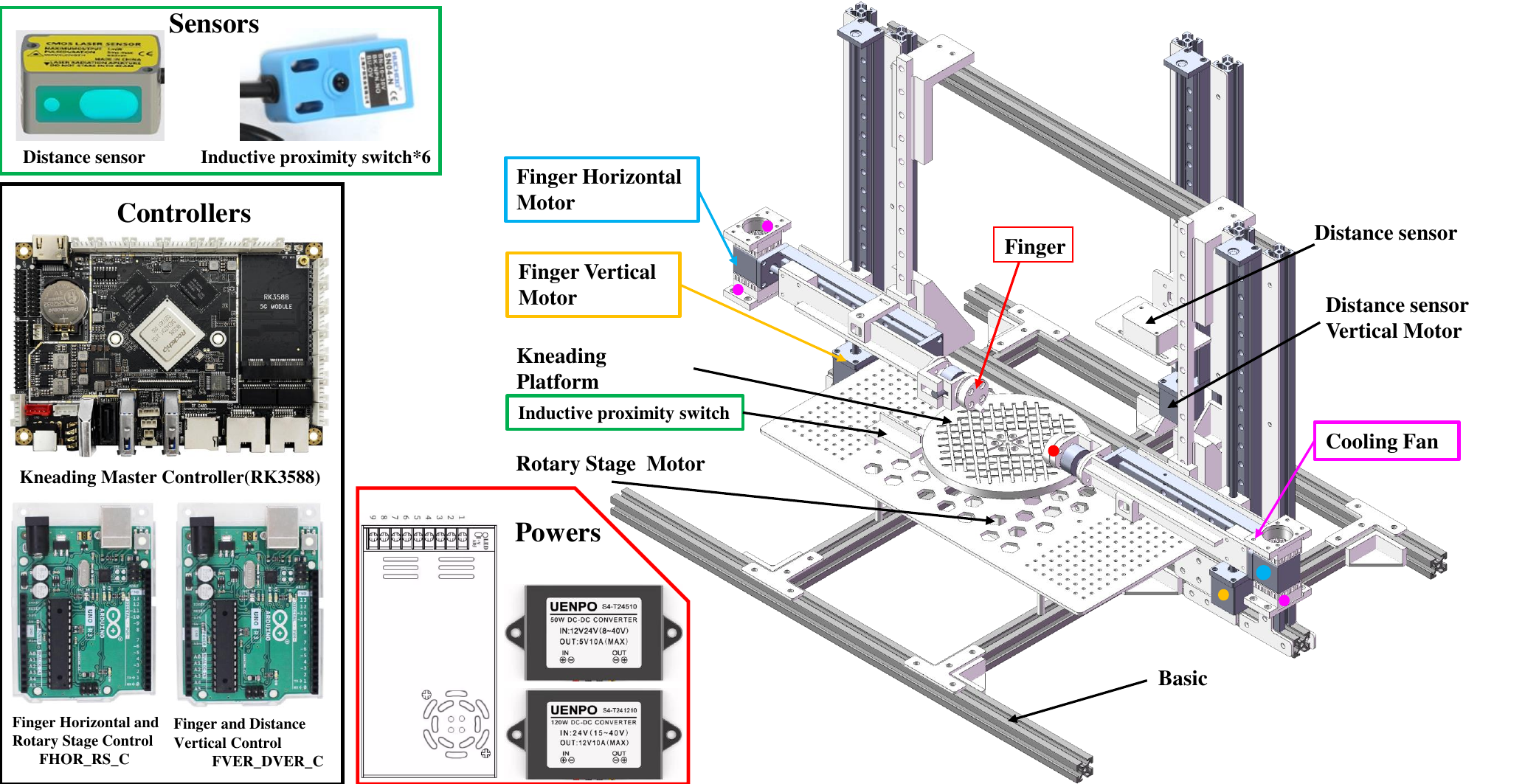}
\caption{\small Hardware structure diagram}
\label{Hardware structure diagram}
\end{figure}

The lower-level control system is built around an RK3588 platform (ROCKCHIP, Fuzhou, Fujian, China), 
equipped with a 128\,GB solid-state drive (Yangtze Memory Technologies, Wuhan, Hubei, China) 
and 8\,GB of RAM (Samsung Electronics, Suwon, Gyeonggi-do, South Korea), 
and running an Ubuntu~20.04 development environment. 
The RK3588 communicates with a laser distance sensor (BOJKE, Shenzhen, Guangdong, China), 
which provides a measurement range of $80\,\mathrm{mm}$, a measurement accuracy of $0.2\,\mathrm{mm}$, 
and a maximum data acquisition frequency of $100\,\mathrm{Hz}$, via the RS485 protocol 
(MAXIM, California, United States). 
In addition, the RK3588 interfaces with two Arduino Uno boards (Arduino AG, Ivrea, Italy) 
through serial and RS232 connections, respectively. 
The two Arduino boards run an improved version of the Arduino CNC firmware and are responsible for 
controlling five 28-stepper motors (SUHENG, YanCheng, JiangSu, China) 
with a step angle of \ang{1.8}, a rated current of $0.8\,\mathrm{A}$, 
and a holding torque of $0.08\,\mathrm{N\cdot m}$, 
as well as one 57-stepper motor (YINGPENG, Shenzhen, Guangdong, China) 
with a step angle of \ang{1.8}, a rated current of $4.5\,\mathrm{A}$, 
and a holding torque of $0.5\,\mathrm{N\cdot m}$. 
Position and state feedback are provided by six inductive proximity switches 
(SUHENG, YanCheng, JiangSu, China), which output digital signals in \texttt{1/0} mode. 
Four cooling fans (YIHE, PuNing, Guangdong, China) are installed to dissipate heat 
from the two horizontally moving 28-stepper motors. 
The entire system is powered by a DC switching power supply 
that converts $220\,\mathrm{VAC}$ to $24\,\mathrm{V}$ at $20\,\mathrm{A}$ 
(HONGMING, Zhongshan, Guangdong, China). 
The overall hardware architecture of the system is illustrated in Fig.~\ref{Hardware structure diagram}.

\begin{figure}[H]
\centering
\includegraphics[width=1.0\columnwidth]{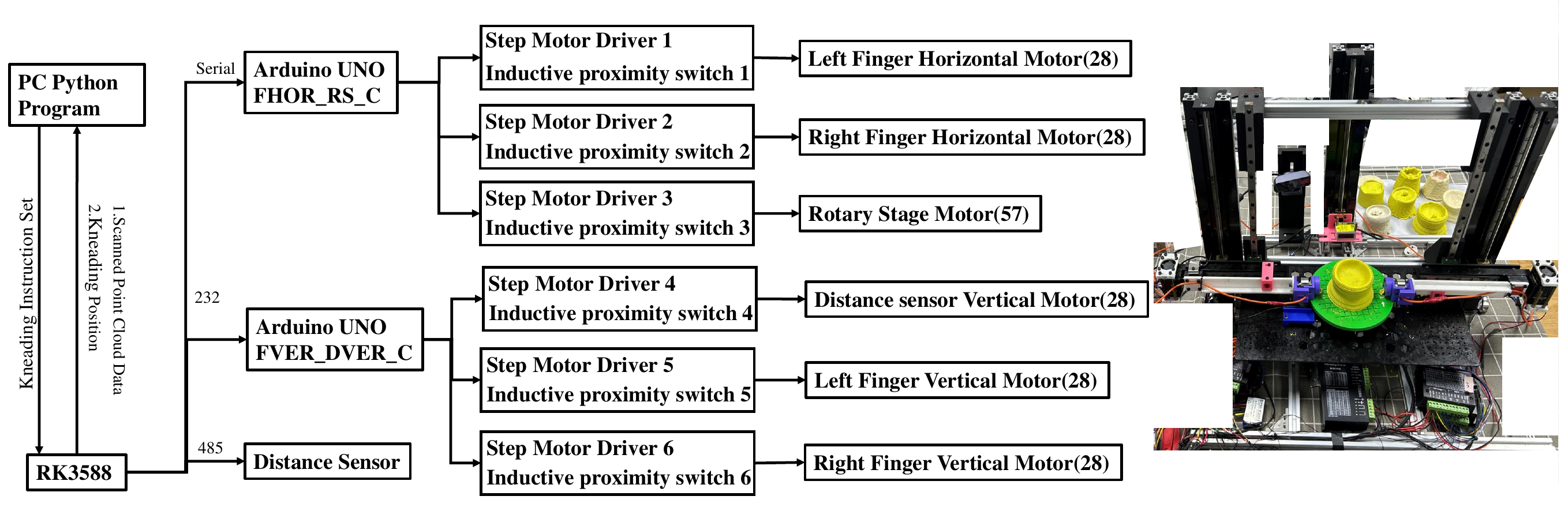}
\caption{\small Control and sensor system relationship diagram}
\label{Control and sensor system}
\end{figure}

Main relationship and physical components of the kneading sensor and control system in Fig.~\ref{Control and sensor system}. 
The RK3588 controls the $FHOR\_RS\_C$ and $FVER\_DVER\_C$ based on the “Kneading Instruction Set” received from the upper computer, using G-code commands. The $FHOR\_RS\_C$ is responsible for controlling the horizontal movement of the left and right fingers as well as the rotation of the rotary stage, while the $FVER\_DVER\_C$ controls the vertical motion of the two fingers and the distance sensor. 

\begin{figure}[H]
\centering
\includegraphics[width=1.0\columnwidth]{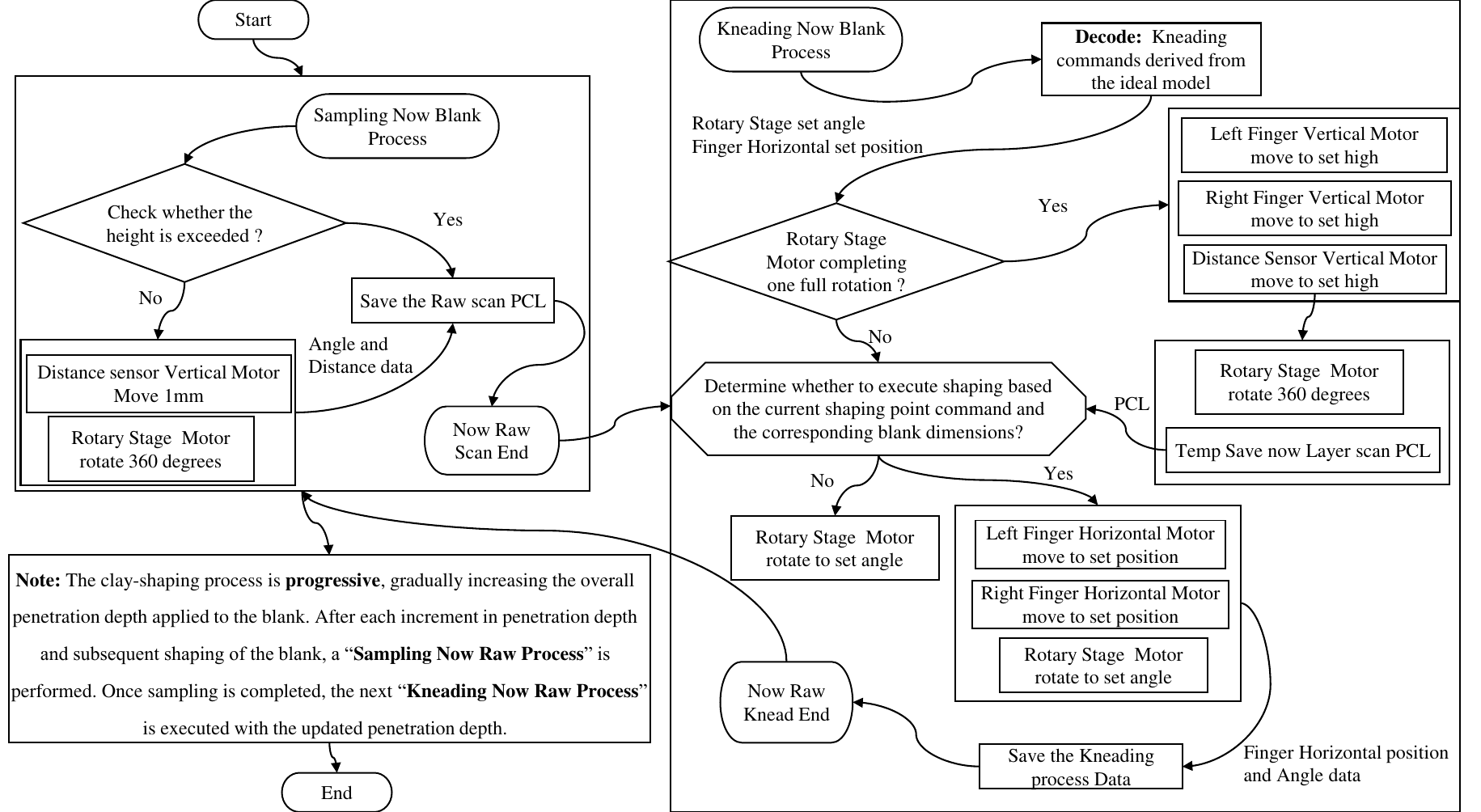}
\caption{\small Control flowchart}
\label{Control flowchart}
\end{figure}

\par During the kneading process, the $FHOR\_RS\_C$ uploads the horizontal position data of the fingers to the RK3588, which stores the data. After completing the kneading operation for the current layer, the $FVER\_DVER\_C$ moves the fingers and the sensor upward to perform kneading on the next layer. After each complete kneading operation corresponding to the preset feeding depth, the distance sensor performs a full scan of the kneaded object. Working together with the $FHOR\_RS\_C$, the system records the rotary stage motor angle and distance sensor data, forming a point cloud dataset. The complete control flow is illustrated in Fig.~\ref{Control flowchart}.

\subsection{Design of the 3D Model for Kneading Geometry}
\begin{figure}[H]
\centering
\begin{adjustwidth}{-1cm}{-1cm} 
    \includegraphics[width=1.1\columnwidth]{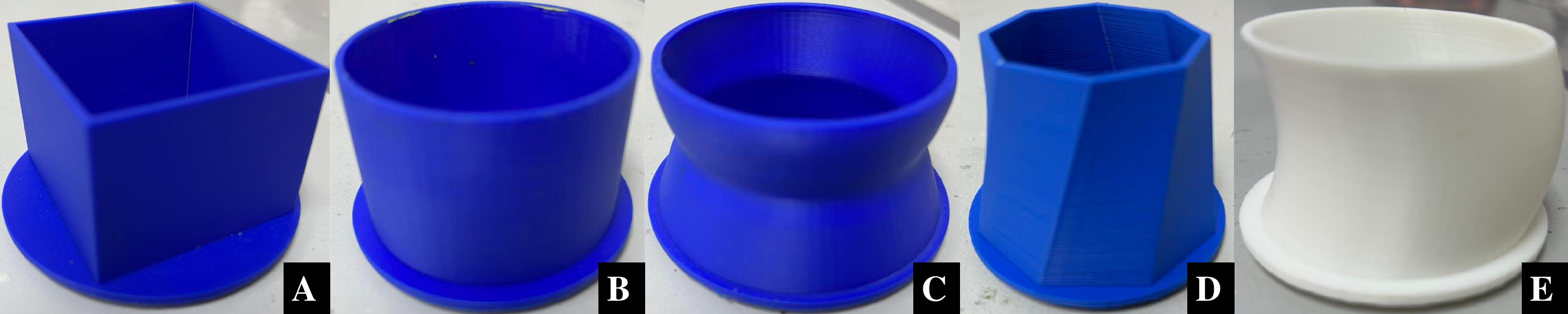}
\end{adjustwidth}
\caption{\small The 3D-printed geometry}
\label{3D_Print_picture}
\end{figure}

Five geometric models were designed, including a square prism with a base side length of 53 mm and a height of 40 mm, a cylinder with a diameter of 60 mm and a height of 40 mm, as well as three geometrically complex shapes: the Slim Waist, Helical Truncated Octagonal Frustum, and Concave Cylinder. The geometric parameters of these three complex geometries are provided in Sections~\ref{sec:Slim waist math equation},~\ref{sec:Helical truncated octagonal frustum math equation} and ~\ref{sec:Concave cylinder math equation}. Fig.~\ref{3D_Print_picture} show the geometries modeled in SolidWorks and the corresponding physical objects fabricated using a Bambu Lab 3D printer. 

\subsubsection{Slim waist math equation}
\label{sec:Slim waist math equation}
\begin{figure}[H]
\centering
    \includegraphics[width=0.9\columnwidth]{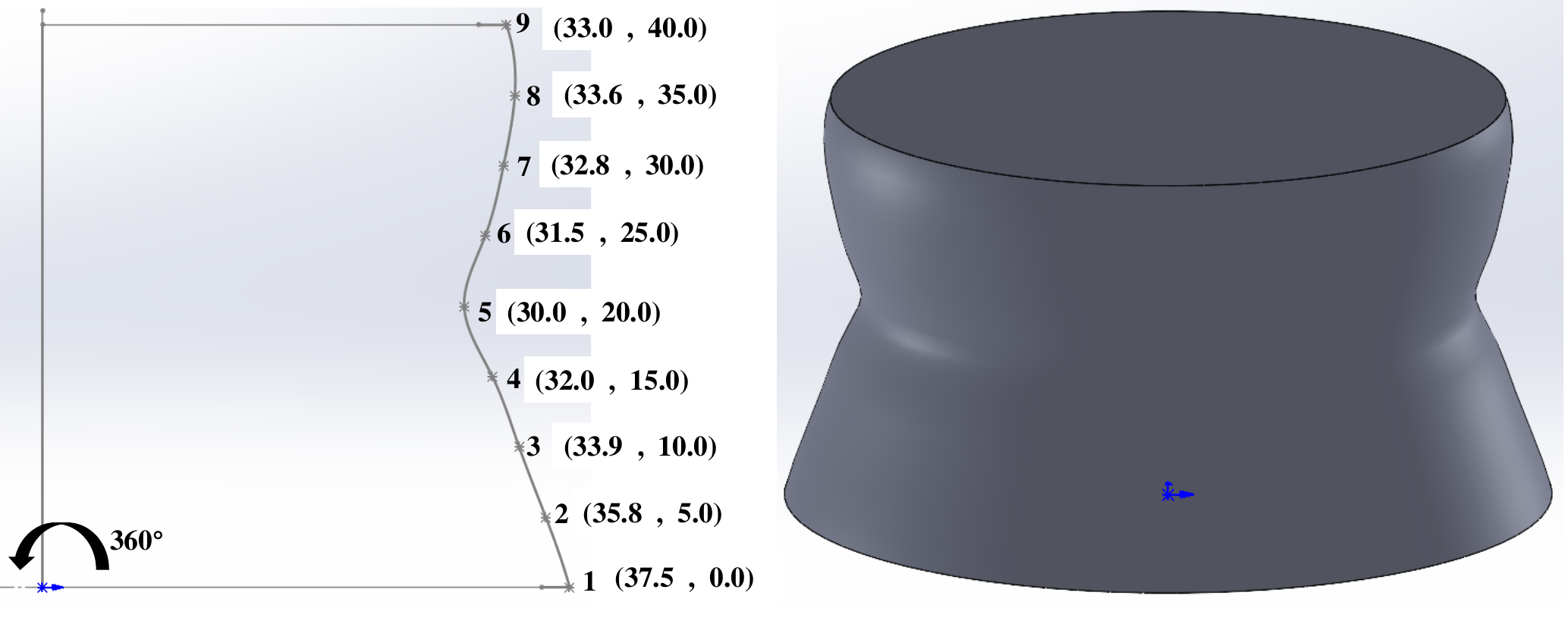}
\caption{\small Constraint points and 3D model of the slim waist curve}
\label{Slim waist params and model}
\end{figure}

Slim waist the curve is generated using the spline curve feature in SolidWorks software, which is a B-spline curve. A two-dimensional cubic B-spline is used. The formula is as follows: let the control points be \(P_i = (x_i, y_i), \quad i = 0, \dots, 8,\)and the degree be \(p = 3\),  \(P_i = (x_i, y_i)\) and 3D model in Fig~.\ref{Slim waist params and model}. Let \(C(t)\) denote the point on the spline curve corresponding to the parameter \(t\), where \(N_{i,3}(t)\) is the cubic B-spline basis function. The parametric form of the B-spline is Equ.\ref{The basic formula of cubic B-spline curves}, Here, \(N_{i,3}(t)\) denotes the cubic B-spline basis function, defined recursively by the Cox--de Boor formula. \(N_{i,0}(t)\) is the zero-degree basis function is Equ.\ref{B-spline Zero Function}, and \(N_{i,p}(t)\) represents the B-spline basis function of degree \(p\) associated with the \(i\)-th control point \(P_i\) is Equ.\ref{B-spline Normal Function}. In these expressions, \(t\) is the curve parameter, \(i\) is the control point index, \(p\) is the degree of the basis function, and \(u_i\) denotes the \(i\)-th knot.

\begin{equation}
C(t) = \sum_{i=0}^{8} N_{i,3}(t)\,P_i, \quad t \in [0,1]
\label{The basic formula of cubic B-spline curves}
\end{equation}

\begin{equation}
N_{i,0}(t) =
\begin{cases}
1, & \text{if } u_i \le t < u_{i+1} \\
0, & \text{otherwise}
\end{cases} 
\label{B-spline Zero Function}
\end{equation}

\begin{equation}
N_{i,p}(t) =
\frac{t - u_i}{u_{i+p} - u_i} \, N_{i,p-1}(t) +
\frac{u_{i+p+1} - t}{u_{i+p+1} - u_{i+1}} \, N_{i+1,p-1}(t)
\label{B-spline Normal Function}
\end{equation}

The curve generated by the B-spline is revolved around the \(Z\)axis to create the slim-waist surface, where the radius of the surface at each point corresponds to the points on the B-spline curve is Equ.\ref{Slim waist Radius function}, the \(R_i\) is each control point. Parametric equation of the slim-waist surface is Equ.\ref{Slim waist Parametric equation}.

\begin{equation}
r(z) = \sum_{i=0}^{n} N_{i,3}\bigl(t(z)\bigr)\, R_i,
t(z) = \frac{z - z_{\min}}{z_{\max} - z_{\min}}
\label{Slim waist Radius function}
\end{equation}

\begin{equation}
X(\theta, z) = (x, y, z) = \bigl( r(z)\cos\theta,\; r(z)\sin\theta,\; z \bigr),
\quad \theta \in [0, 2\pi),\; z \in [0, h]
\label{Slim waist Parametric equation}
\end{equation}

\subsubsection{Helical truncated octagonal frustum math equation}
\label{sec:Helical truncated octagonal frustum math equation}
\begin{figure}[H]
\centering
    \includegraphics[width=0.8\columnwidth]{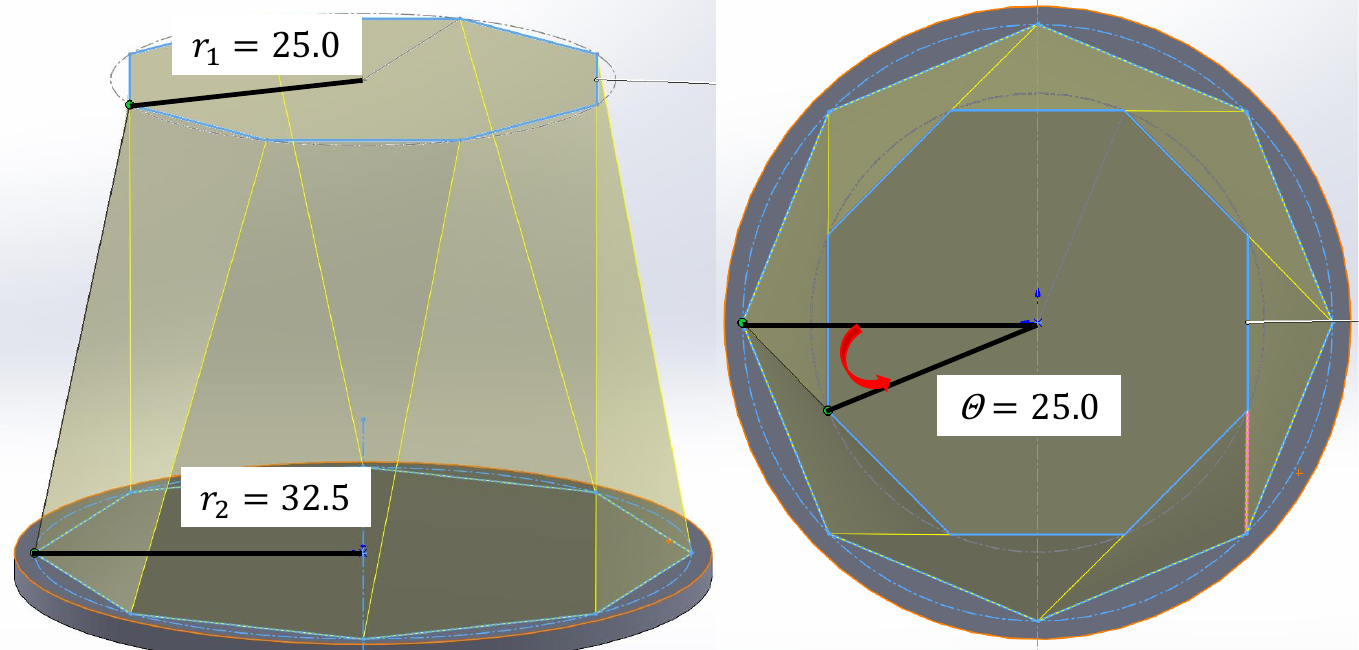}
\caption{\small Helical truncated octagonal frustum params and model}
\label{Helical octagonal frustum params}
\end{figure}

Let the total height be \(h\), the number of layers be \(N\), the inradius of the upper inscribed octagon be \(r_{1}\), the inradius of the lower inscribed octagon be \(r_{2}\), and the total twist angle be \(\Theta\), in Fig~.\ref{Helical octagonal frustum params} (i.e., the rotation of the lower base relative to the upper base) . For the \(k\)-th layer, the parameters are defined as follows: the parameter \(t\) is a normalized variable (typically mapped along the height direction, either \(0 \sim 1\) or \(0 \sim N\) layers); \(r(t)\) is the radius function that determines the cross-sectional size of the layer and can represent either a linear frustum or a curved variation; \(z(t)\) is the height function that gives the coordinate along the \(Z\)-axis; \(\theta(t)\) is the accumulated twist angle along the height that governs the helical or torsional behavior, is Equ.\ref{Helical truncated octagonal frustum basic function}. And \(\phi_i\) is the initial angular position of the \(i\)-th vertex before twisting, the \(k\) layers i nodes is Equ.\ref{Helical truncated octagonal frustum point function}

\begin{equation}
\begin{cases}
t = \frac{N}{k}, \\
r(t) = r_1 + (r_2 - r_1) t, \\ 
\theta(t) = \Theta t, \; z(t) = h t
\end{cases}
\label{Helical truncated octagonal frustum basic function}
\end{equation}

\begin{equation}
\begin{cases}
\phi_i = \frac{8 \pi}{2} i, \\
P_{k,i} = \big( r(t) \cos(\phi_i + \theta(t)),r(t) \sin(\phi_i + \theta(t)),\; z(t) \big)
\end{cases}
\label{Helical truncated octagonal frustum point function}
\end{equation}

\subsubsection{Concave cylinder math equation}
\label{sec:Concave cylinder math equation}
\begin{figure}[H]
\centering
\includegraphics[width=0.5\columnwidth]{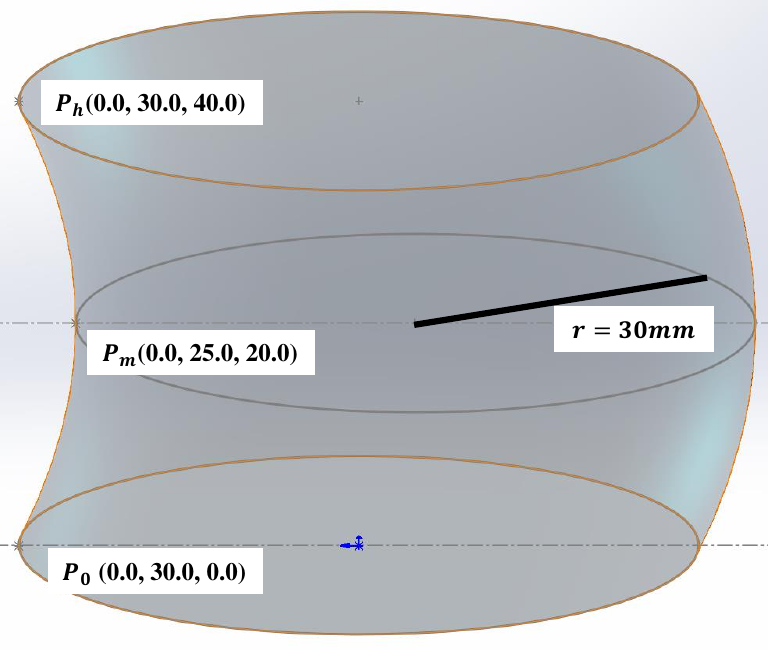}
\caption{\small Concave cylinder params and model}
\label{Concave cylinder params}
\end{figure}

Concave cylinder use SolidWorks drawing, the surface is generated by lofting through three points:\(P_{0} = (0,\, r,\, 0), \quad P_{m} = (0,\, y_{0} + r,\, \tfrac{h}{2}), \quad P_{h} = (0,\, r,\, h)\), in Fig~.\ref{Concave cylinder params}. Let the center of the circle be denoted by \(C\), and its trajectory by \(C(z)\) is Equ.\ref{Concave cylinder center trajectory}, \(d(z)\) is Equ.\ref{Concave cylinder center in YOZ}. The parametric formulation of a concave cylinder is Equ.\ref{Concave cylinder parametric formulation}

\begin{equation}
C(z) = (0, d(z), z), \; z \in [0, h]
\label{Concave cylinder center trajectory}
\end{equation}

\begin{equation}
d(z) = 4y \cdot \frac{z}{h} \left( 1 - \frac{z}{h} \right)
\label{Concave cylinder center in YOZ}
\end{equation}

\begin{equation}
\mathbf{X}(\theta, z) = \bigl(x(\theta, z),\, y(\theta, z),\, z\bigr) 
= \bigl(r \cos \theta,\, d(z) + r \sin \theta,\, z\bigr)
\label{Concave cylinder parametric formulation}
\end{equation}

\section{Geometric Point Cloud Processing for Kneading Generation}

\subsection{Layer-Based Contour Extraction Algorithm from STL Mesh with Convex Hull}
The STL mesh model consists of a large number of triangular facets that describe the surface geometry of a 3D object. To extract cross-sectional contours for subsequent analysis or manufacturing simulation, a layer-based contour extraction algorithm is applied. The algorithm slices the STL model along the vertical \(Z\) direction with a fixed layer thickness of \(\Delta h = 0.1~\text{mm}\). For each slicing plane at height \(z = h_i\), the intersection between the plane and the mesh is computed. For a triangular facet with vertices \(P_1(x_1, y_1, z_1)\), \(P_2(x_2, y_2, z_2)\), \(P_3(x_3, y_3, z_3)\), the plane intersects the triangle if satisfies Equ.\ref{Plane intersects the Triangle}. 

\begin{equation}
\begin{cases}
(z_1 - h_i)(z_2 - h_i) < 0 \quad \text{or} \quad \\
(z_2 - h_i)(z_3 - h_i) < 0 \quad \text{or} \quad \\
(z_3 - h_i)(z_1 - h_i) < 0
\end{cases}
\label{Plane intersects the Triangle}
\end{equation}

\par
\begin{figure}[H]
\centering
\includegraphics[width=0.8\columnwidth]{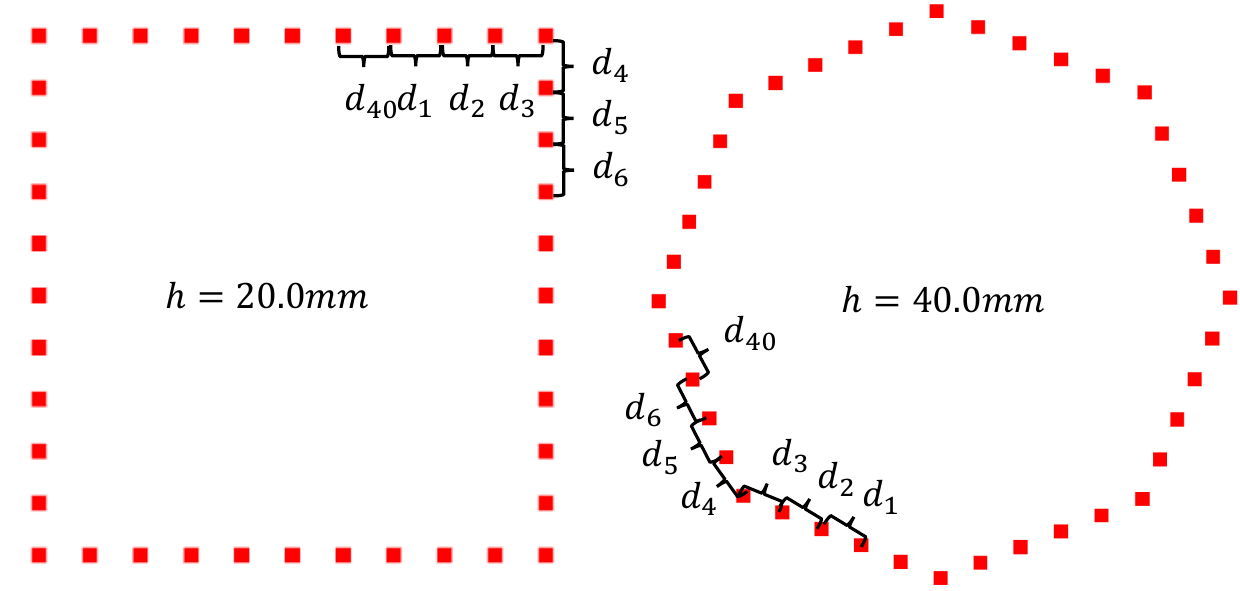}
\caption{\small Single layer point cloud sampling}
\label{Single layer point cloud sampling}
\end{figure}

 The intersection points along the edges are obtained by linear interpolation as Equ.\ref{Linear interpolation}, and similarly for other edges. All intersection points of the current layer form the contour point set \(\mathcal{P}_i = \{ P_{i1}, P_{i2}, \dots, P_{in} \}\). To obtain a smooth, ordered outer contour, the algorithm applies a \textbf{convex hull} operation. Let \(\mathcal{P}_i \subset \mathbb{R}^2\) denote the 2D projection of the contour points onto the XY plane. The convex hull is defined as the minimal convex polygon enclosing all points as Equ.\ref{ConvexHull equation}. Once \(\mathcal{C}_i\) is obtained, the perimeter of the contour \(L_i\) is calculated as Equ.\ref{Layer points cloud contour}, and the contour is resampled uniformly into \(N = 400\) points with spacing \(d\) as Equ.\ref{Point and point distance}. This procedure is repeated for all slicing planes as in Fig~.\ref{Single layer point cloud sampling}, \(d_1=d_2=d_3=......=d_{40}\)(in Figure sampling numbers is 40, use A and D because the cross-sections B, C, and E are all circular), generating a series of evenly spaced, ordered contour layers, which serve as the basis for geometric analysis, surface reconstruction, or toolpath generation in manufacturing processes.

\begin{equation}
\begin{cases}
P_{12} = P_1 + \frac{h_i - z_1}{z_2 - z_1} (P_2 - P_1) \\
P_{13} = P_1 + \frac{h_i - z_1}{z_3 - z_1} (P_3 - P_1)
\end{cases}
\label{Linear interpolation}
\end{equation}

\begin{equation}
\mathcal{C}_i = \text{ConvexHull}(\mathcal{P}_i) = 
\left\{ \sum_{k=1}^{m} \lambda_k P_k \ \Big| \ \lambda_k \ge 0, \ \sum_{k=1}^{m} \lambda_k = 1 \right\}_{\text{boundary}}.
\label{ConvexHull equation}
\end{equation}

\begin{equation}
L_i = \sum_{k=1}^{n-1} \| P_i(k+1) - P_{ik} \| + \| P_{i1} - P_{in} \|
\label{Layer points cloud contour}
\end{equation}

\begin{equation}
d = \frac{L_i}{N}
\label{Point and point distance}
\end{equation}

\subsection{Determination of Whether a Geometry is an Enveloping Type}
\label{Generate kneading commands based on geometric characteristics}
According to whether the geometry is of the enveloping type, different kneading methods are applied to geometries such as the cylinder, quadrangular prism, helical truncated octagonal frustum, Slim waist, and concave cylinder. The geometric body is considered an enveloping shape when it simultaneously satisfies the following three criteria: continuity of average radius gradient as Equ.\ref{Continuity of average radius gradient}, monotonicity of cross-sectional area as Equ.\ref{Monotonicity of cross-sectional area}, and regularity of torsion angle as Equ.\ref{Regularity of torsion angle}. While geometry satisfy \(\frac{d\bar{r}}{dz}\), \(\frac{dA}{dz}\) and \( f(z)\) is continuous, the geometry is considered an enveloping geometry. 

\begin{equation}
\bar{r}(z) = \frac{1}{2\pi} \int_0^{2\pi} r(z,\theta) \, d\theta, \quad
\frac{d\bar{r}}{dz}
\label{Continuity of average radius gradient}
\end{equation}

\begin{equation}
 A(z) = \int_0^{2\pi} r^2(z,\theta) \, d\theta, \quad
        \frac{dA}{dz}
\label{Monotonicity of cross-sectional area}
\end{equation}

\begin{equation}
\frac{d\theta}{dz} = f(z),
\label{Regularity of torsion angle}
\end{equation}

\begin{table}[H]
\centering
\caption{Geometry Classification}
\resizebox{\textwidth}{!}{
\begin{tabularx}{\textwidth}{c|X|X|X|X}
\hline
\textbf{Index} & \textbf{Variation of Circumscribed Radius} & \textbf{Section Center Variation} & \textbf{Symmetry Property} & \textbf{Enveloping Shape} \\ 
\hline
A & Constant & Fixed & Axial, Central Symmetry & True \\ 
B & Constant & Coaxial & Axial, Central, and Rotational Symmetry & True \\ 
C & Continuous variation & Coaxial & Rotational Symmetry & False \\ 
D & Linear Variation & Fixed Axis & Rotational Symmetry along Axis & True \\ 
E & Constant & Center Translates on Plane along $z$ & Axial Symmetry & False \\ 
\hline
\end{tabularx}
}
\label{Geometry classied table}
\end{table}

\par The circumscribed radius of the sections, section centers, rotational/symmetry properties, and enveloping status of the five geometric bodies are shown in the Tab.\ref{Geometry classied table}.

\subsubsection{Envelope shaping first}
\label{Envelope shaping first method kneading commands}
The point cloud data are organized into layers according to height, and each layer contains \(N_h = 400\) points as Equ.\ref{Points clouds axis}. Convert the point \((x_{h,i}, y_{h,i})\) into polar coordinates as Equ.\ref{Points cloud axis to polar}, where \(r_{h,i} = D_{h,i}\). A symmetric pairing is established between the left and right fingers as Equ.\ref{Left Right Finger Set pos}, \(\{D_{h i}\}_{i=1}^{N_h'} = L_h \cup R_h\). During the kneading process, the turntable rotates over a full circle, while Equ.\ref{Left Right Finger Set pos} only represents half of the rotation. For the other half, the point cloud data of the left and right fingers are swapped.

\begin{equation}
P = \bigcup_{h = 1}^{H_{\max}} P_h, \quad 
\text{where } 
P_h = \{ (x_{h,i},\, y_{h,i},\, z_{h,i}) \mid z_{h,i} = h,\ i = 1,\dots, N_h \}.
\label{Points clouds axis}
\end{equation}

\begin{equation}
\begin{cases}
r_{hi} = \sqrt{x_{hi}^2 + y_{hi}^2} = D_{hi} \\
\theta_{hi} = \operatorname{atan2}(y_{hi}, x_{hi}) \in (-\pi, \pi] \\
z_{hi} = h
\end{cases}
\label{Points cloud axis to polar}
\end{equation}

\begin{equation}
\begin{cases}
L_h = \{D_{h1}, \dots, D_{h,\frac{N_h'}{2}}\}
\\
R_h = \{D_{h,\frac{N_h'}{2}+1}, \dots, D_{h,N_h'}\}
\end{cases}
\label{Left Right Finger Set pos}
\end{equation}

\par The kneading–forming process is characterized by a layer-by-layer gradual transition. To prevent over-compression or under-forming during shaping, a molding compensation term is introduced into the end-effector displacement as a global correction parameter, as Equ.\ref{Mold Raw Set pos calculate}, the \(W\) is left and right finger in ideal state width. The ideal volume of the molded object is calculated as Equ.\ref{Origin volume calculate}, \(h_k\) is the max radius of \(k\) layers, The compensated volume is given by Equ.\ref{Compensate volume calculate}, \(D'(h_k)\) represents the maximum radius of the current \(k\)-th layer following the compensation.

\begin{equation}
D_{h i}^{\text{raw}} = 
\frac{W}{2} - \bigl|D_{h i} - s\bigr| - m ,
\quad\text{where } m = \text{mold\_scale}.
\label{Mold Raw Set pos calculate}
\end{equation}

\begin{equation}
V_{\text{orig}} = \sum_{k=1}^{n} \pi \, D(h_k)^2 \cdot \Delta h, 
\label{Origin volume calculate}
\end{equation}

\begin{equation}
\begin{cases}
D'(h_k) = D(h_k) + \Delta r \\
V_{\text{scaled}} = \sum_{k=1}^{n} \pi \, D'(h_k)^2 \cdot \Delta h
\end{cases}
\label{Compensate volume calculate}
\end{equation}

\par To improve the kneading efficiency, the radius is increased while keeping the volume constant, and the height is recalculated according to the volume ratio. First calculated origin volume and scaled volume ration as Equ.\ref{Origin and Scaled ration calculated}, \(H'\) denotes the height to be kneaded as Equ.\ref{Actual Mold High Calculate}. By calculating the circumference of the point cloud \(C_h\) as Equ.\ref{Raw circumference calculate} in the current layer after compensation and dividing it by the diameter of the end-effector \(d_{m}\), the number of kneading operations required for the current layer is obtained as Equ.\ref{Knead numbers calculate}. The kneading step size is then determined based on this number of operations as Equ.\ref{Knead numbers calculate}, while \(m\) zero is the final forming stage of kneading, the value of \(K_h\) is calculated base on \(d_{m}/2\). In the last kneading command is \((h,\, i,\, D_{hi}^{\text{raw}},\, i + N_h'/2,\, D_{h,\,i+N_h'/2}^{\text{raw}})\).

\begin{equation}
\alpha = \dfrac{V_{\text{orig}}}{V_{\text{scaled}}} \\
\label{Origin and Scaled ration calculated}
\end{equation}

\begin{equation}
H' = \{\,h'_1, h'_2, \dots, h'_n \mid 
h'_k = h_1 + \alpha \,(h_k - h_1)\}.
\label{Actual Mold High Calculate}
\end{equation}

\begin{equation}
C_h = 2\pi D'(h_k)
\label{Raw circumference calculate}
\end{equation}

\begin{equation}
\begin{cases}
K_h = \left\lceil \frac{C_h}{d_{m}} \right\rceil \\
K_h = \left\lceil \frac{C_h}{d_{m}/2} \right\rceil , \quad while \quad m=0 
\end{cases}
\label{Knead numbers calculate}
\end{equation}

\begin{equation}
s_h = \max\left(1,\; \left\lfloor \frac{N_h'}{K_h} \right\rfloor \right)
\label{Knead step Calculate}
\end{equation}

\subsubsection{Similar gradient method}
\label{Similar gradient method kneading commands}
The similar gradient method is a manufacturing approach based on maintaining local consistency of shape across adjacent layers. Its core principle is to achieve global continuity and controllability of deformation through gradual variation between neighboring points or cross-sections. Calculate the cross-sectional area of each layer of the point cloud based on its polar coordinate data as Equ.\ref{Calculate single layer area}, then sum the every layers area as Equ.\ref{Similar Volume calculate}, compressed ration calculate as \ref{Origin and Scaled ration calculated}. Compressed \(H_{\max}^{\text{compressed}}\) calculate as Equ.\ref{Compressed max high calculate}, map the points of the compressed layer \(H_k^{\text{compressed}}\) to the nearest original layer \(H^{\text{scaled}}\).

\begin{equation}
\begin{cases}
A_h &= \frac{1}{2} \sum_{i=1}^{n_h} r_i r_{i+1} \sin(\theta_{i+1} - \theta_i), 
\quad r_1 = r_{n_h+1}, \theta_1 = \theta_{n_h+1} \\[1mm]
A_h^{\text{scaled}} &= \frac{1}{2} \sum_{i=1}^{n_h} r_i^{\text{new}} r_{i+1}^{\text{new}} \sin(\theta_{i+1} - \theta_i) \\[2mm]
\end{cases}
\label{Calculate single layer area}
\end{equation}

\begin{equation}
\begin{cases}
V_{\text{orig}} = \sum_{h} A_h \\
V_{\text{scaled}} = \sum_{h} A_h^{\text{scaled}} \\[1mm]
\end{cases}
\label{Similar Volume calculate}
\end{equation}

\begin{equation}
H_{\max}^{\text{compressed}} = H_{\min} + (H_{\max} - H_{\min}) \cdot \alpha \\[2mm]
\label{Compressed max high calculate}
\end{equation}

\begin{equation}
H_k^{\text{compressed}} = H_{\min} + k \cdot \Delta H, \quad
k = 0,1,\dots,N-1, \quad
H_{N-1}^{\text{compressed}} \le H_{\max}^{\text{compressed}} \\[2mm]
\label{Compressed layers high calculated}
\end{equation}

\begin{equation}
\begin{cases}
\text{idx} = \left\lfloor \frac{k}{N} \cdot N_{\text{layers}} \right\rfloor, \quad
\text{layer\_H}[\text{idx}] = H^{\text{scaled}}\\[1mm]
\{(H_k^{\text{compressed}}, A_i, D_i^{\text{raw}})\}_{i=1}^{n_{\text{layer}}} 
\end{cases}
\end{equation}

\subsection{From Point Cloud to Kneading Command: Overview}
\label{Summary of molding command generation}
The proposed kneading instruction generation algorithm first extracts point cloud data from the STL mesh model through layer-based slicing and converts it into a polar coordinate representation for geometric analysis. Based on the continuity of the average radius gradient, the monotonicity of the cross-sectional area, and the regularity of the torsion angle, the geometry is classified as either enveloping or non-enveloping. For enveloping geometries, the Envelope Shaping First method is applied, whereas the Similar Gradient Method is used for non-enveloping geometries.

\par Subsequently, the kneading path is adaptively generated according to the minimum-radius principle and the geometric characteristics of each layer. The algorithm determines the feed and step parameters based on the end-effector diameter and radial compensation, ensuring a continuous and volume-consistent forming process. The final output consists of an optimized kneading instruction sequence for precise and efficient shaping. Algorithm is summarized in Code~\ref{alg:Kneading_generation_pseudocode}

\begin{algorithm}[H]
\caption{Adaptive Kneading Data Generation Based on Geometric Features}
\label{alg:Kneading_generation_pseudocode}
\begin{algorithmic}[1]
\REQUIRE STL mesh file, end-effector diameter $d_m$, blank point cloud
\ENSURE Kneading command set $Kneading\_Commands$

\STATE Read STL file and extract mesh vertices
\STATE Generate point cloud $P = \{p_i(x_i, y_i, z_i)\}$ and sort by height $z$

\FOR{each slicing layer $h_i$}
    \STATE Set layer thickness $\Delta h = 0.1$ mm
    \STATE Extract current layer point cloud $C_i$
    \STATE Convert $C_i$ to polar coordinates $(r_i, \theta_i)$
    \STATE Compute mean radius $\bar{r}_i$, section area $A_i$, and torsion function $f(z_i)$
\ENDFOR

\STATE Determine geometry type:
    \IF{ $\frac{d\bar{r}}{dz}$, $\frac{dA}{dz}$, and $f(z)$ are continuous }
        \STATE Geometry is \textbf{enveloping type}
    \ELSE
        \STATE Geometry is \textbf{non-enveloping type}
    \ENDIF

\STATE Select kneading method:
    \IF{enveloping geometry}
        \STATE Apply \textit{Envelope Shaping First} method
    \ELSE
        \STATE Apply \textit{Similar Gradient Method}
    \ENDIF

\STATE For each layer, compute perimeter:
    \STATE \(
        L_i = \sum_{k=1}^{n-1} \| P_{i,k+1} - P_{i,k} \| + \| P_{i,1} - P_{i,n} \|
    \)
    \STATE Resample each layer to $N = 400$ points

\STATE Generate kneading commands based on minimum radius principle:
    \STATE \(
        \Delta r_{\max} = R_{blank}^{\max} - R_{mold}^{\min}
    \)
    \IF{$\Delta r_{\max} \neq 0$}
        \STATE $height\_step = d_m - 1$, \quad $radial\_step = d_m$
        \WHILE{$\Delta r_{\max} > 0$}
            \STATE Perform kneading with feed depth = 1 mm
            \STATE $\Delta r_{\max} \gets \Delta r_{\max} - 1$
        \ENDWHILE
    \ELSE
        \STATE $height\_step = d_m / 2$, \quad $radial\_step = d_m / 2$
    \ENDIF

\STATE Output final kneading commands $Kneading\_Commands$
\end{algorithmic}
\end{algorithm}

\section{Experimental and Data Comparative Analysis}
\subsection{Area of the surface}
General ring-mesh method: core concept (continuous to discrete). In the ideal case, the surface area of a smooth surface \(S\) is given by Equ.\ref{Area of surface math function}, where \(\textbf{n}\) is the normal vector of the surface. However, for a point cloud, the surface is discrete and lacks a parametric equation, normal vectors, or an analytical expression. 
\par Therefore, we approximate the surface integral using a triangular mesh of the surface is Equ.\ref{Area of surface Accumulation of triangle areas} where \(A_i\) is the area of the \(i\)-th triangle in the mesh and \(M\) is the total number of triangles. According to the point cloud sampling method, where there are 400 points per loop with a vertical spacing of \(\Delta z = 1\,\mathrm{mm}\), the point cloud naturally forms a two-dimensional mesh. Here, \(\textbf{v}\) denotes the vertices used to construct the triangles Equ.\ref{The triangle three points}: let \(k\) denote the layer index along the height direction, and \(j\) denote the angular order of points on each loop, with \(M = 400\) points per loop. The points between every two consecutive layers are connected to form quadrilaterals, which are then divided into two triangles as Equ.\ref{Construct quadrilaterals}. here, \(\mathrm{mod}\) denotes the modulo operation. Let the vertices of a quadrilateral be denoted as Equ.\ref{Quadrilaterals points}. The quadrilateral can then be divided into two triangles, as Equ.\ref{Quadrilaterals points to triangle three points}, triangles area calculates as Equ.\ref{Triangles area}. In the last, the total surface area is calculated as Equ.\ref{Sum triangle to Area}

\begin{equation}
A = \iint_{S} \|\mathbf{n}\| \, dS
\label{Area of surface math function}
\end{equation}

\begin{equation}
A \approx \sum_{\Delta} \text{Area}(\Delta)
\label{Area of surface Accumulation of triangle areas}
\end{equation}

\begin{equation}
\mathbf{v}_{k,j} = (x, y, z), \quad k = 0, \ldots, K, \quad j = 0, \ldots, M-1
\label{The triangle three points}
\end{equation}

\begin{equation}
j^+ = (j+1) \bmod M
\label{Construct quadrilaterals}
\end{equation}

\begin{equation}
Q_{k,j} = [ \mathbf{v}_{k,j}, \mathbf{v}_{k+1,j}, \mathbf{v}_{k+1,j^+}, \mathbf{v}_{k,j^+} ]
\label{Quadrilaterals points}
\end{equation}

\begin{equation}
\Delta = 
\begin{cases}
\Delta_1 = (\mathbf{v}_{k,j}, \mathbf{v}_{k+1,j}, \mathbf{v}_{k+1,j^+}) \\
\Delta_2 = (\mathbf{v}_{k,j}, \mathbf{v}_{k+1,j^+}, \mathbf{v}_{k,j^+})
\end{cases}
\label{Quadrilaterals points to triangle three points}
\end{equation}

\begin{equation}
\text{Area}(T) = \frac{1}{2} \| (B-A) \times (C-A) \|
\label{Triangles area}
\end{equation}

\begin{equation}
\begin{aligned}
A_{\text{surface}} 
\approx 
\sum_{k=0}^{K-1} \sum_{j=0}^{M-1} \Bigg[
&\frac{1}{2} 
\left\| 
(\mathbf{v}_{k+1,j} - \mathbf{v}_{k,j}) 
\times 
(\mathbf{v}_{k+1,j^+} - \mathbf{v}_{k,j}) 
\right\|  \\
&+ \frac{1}{2} 
\left\| 
(\mathbf{v}_{k,j^+} - \mathbf{v}_{k,j}) 
\times 
(\mathbf{v}_{k+1,j^+} - \mathbf{v}_{k,j}) 
\right\|
\Bigg]
\label{Sum triangle to Area}
\end{aligned}
\end{equation}


\subsection{Comparison of Kneaded Point Cloud}
\begin{figure}[H]
\centering
\includegraphics[width=0.8\columnwidth]{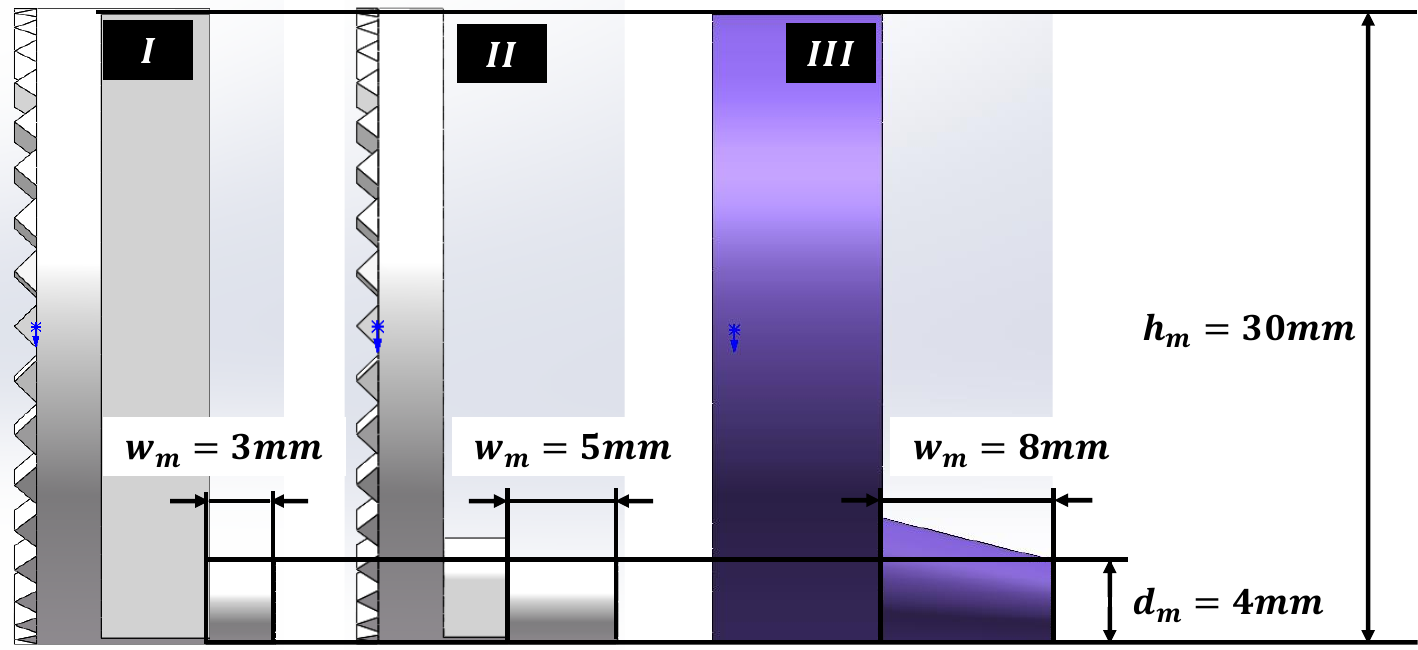}
\caption{\small Three versions of End-effector}
\label{Three versions of End-effector}
\end{figure}

\begin{figure}[H]
\centering
\includegraphics[width=0.8\columnwidth]{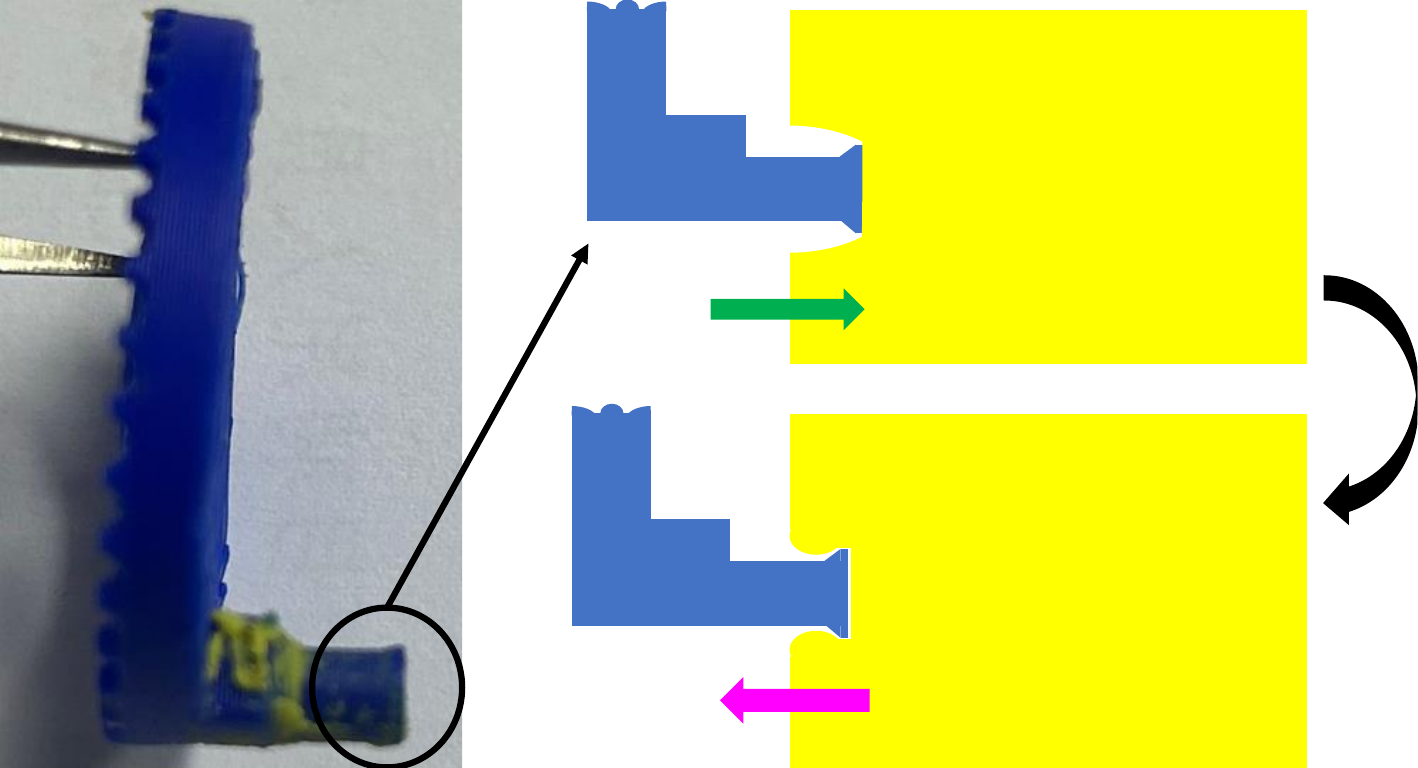}
\caption{\small Retraction pulled out material}
\label{Retraction pulled out material}
\end{figure}

Designed three types of end-effectors, as shown in Fig~.\ref{Three versions of End-effector}. End-effector I (\(w_m\) = 3 mm) pushes the non-kneading surface of the billet when kneading geometries C and E. End-effector II can knead geometries C and E, but after 100,000 kneading cycles, design factors cause burrs to appear on the tool head, which scratch the billet during retraction in Fig~.\ref{Retraction pulled out material}. End-effector III overcomes the shortcomings of I and II, avoiding contact with the non-kneading surface and preventing material from being drawn out during tool retraction thanks to its frustum-shaped head design.

\subsubsection{Ideal Kneading}

\begin{figure}[H]
\centering
\includegraphics[width=0.8\columnwidth]{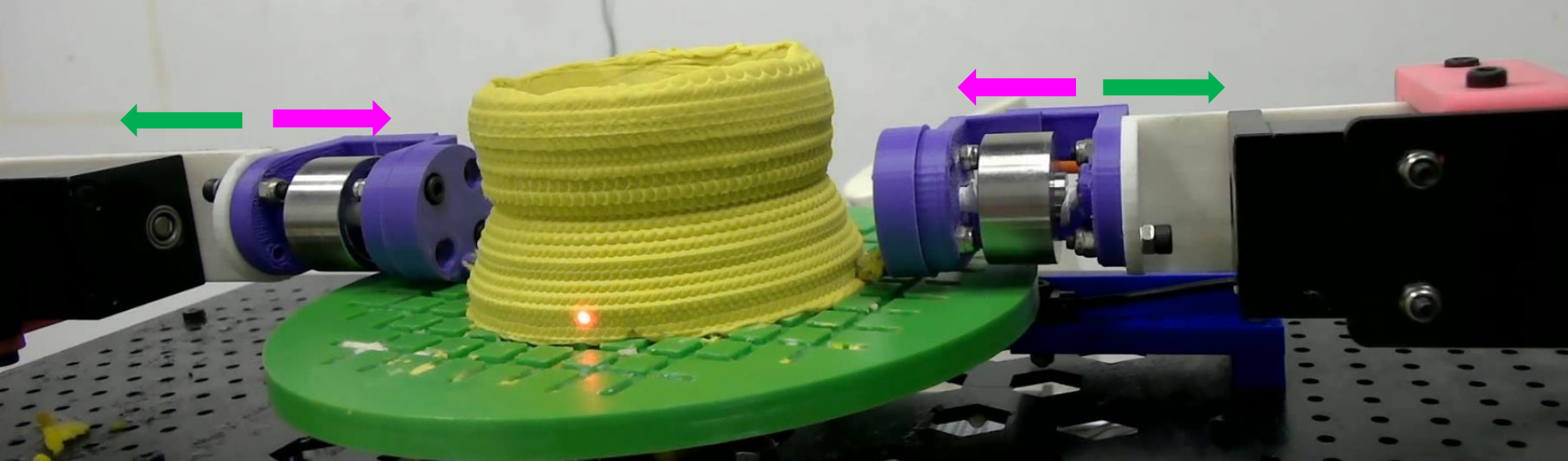}
\caption{\small End-effector reciprocating motion}
\label{End-effector reciprocating motion}
\end{figure}

The end-effector diameter is \(d_m=4mm\). The circular surfaces of two end-effectors are perpendicular to XOY plane and parallel to XOZ plane, moving horizontally toward each other along the Y-axis, as  Fig~.\ref{End-effector reciprocating motion}. 

\begin{figure}[H]
\centering
\includegraphics[width=1.0\columnwidth]{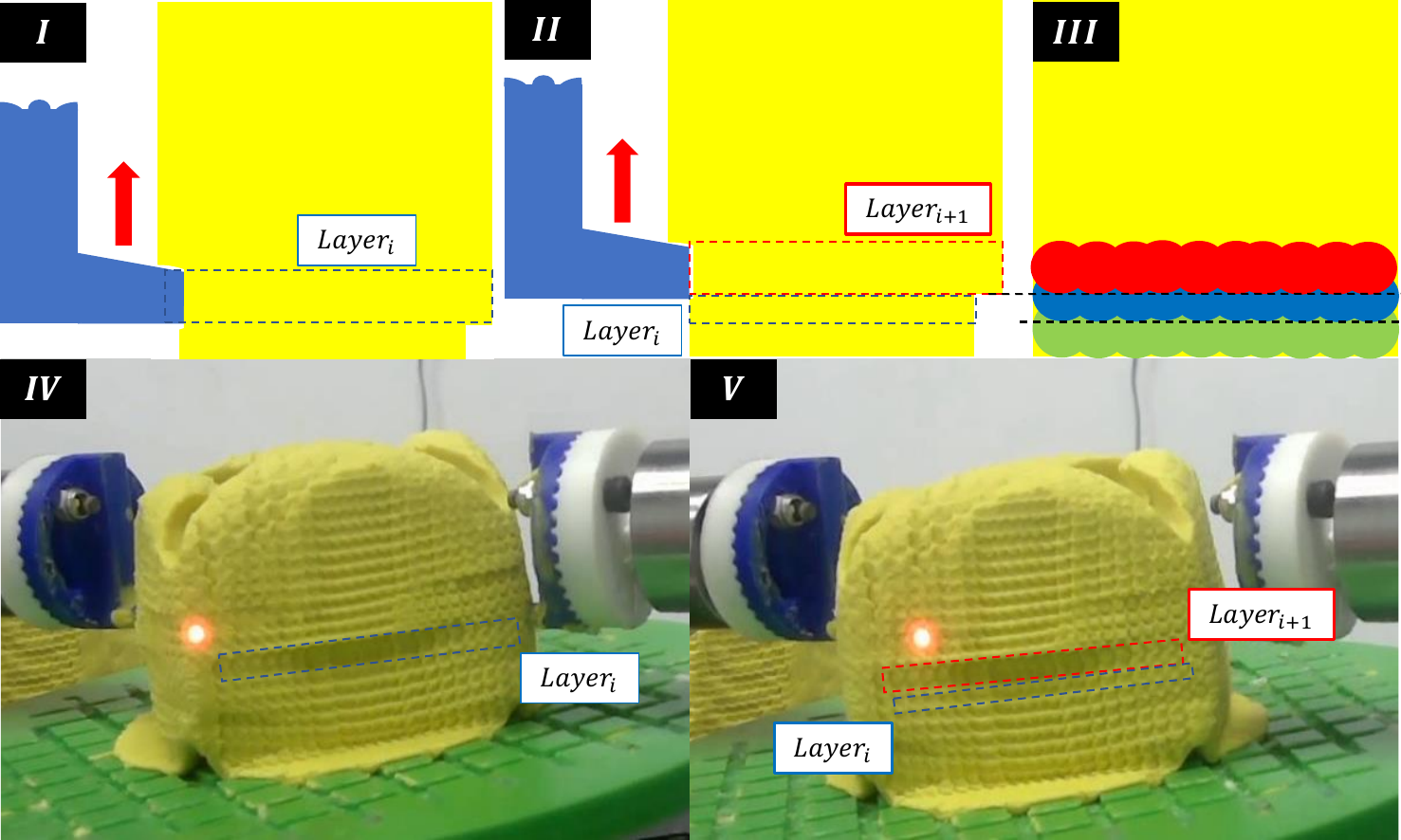}
\caption{\small Actual Knead range}
\label{Actual Knead range}
\end{figure}

kneading is performed layer by layer, and the actual kneading range for each layer is shown in Fig~.\ref{Actual Knead range}. I shows the kneading range of the end-effector at \(layer_i\). After the kneading of \(layer_i\) is completed, \(layer_{i+1}\) is molded as shown in II. III illustrates the effective point cloud ranges of \(layer_i\) and \(layer_{i+1}\). When the molding of \(layer_{i+1}\) is completed, only the lower half of the end-effector’s molding range in \(layer_i\) remains effective (since the molding is performed layer by layer from bottom to top, only the lower half is active). 

Therefore, according to the situation shown in Fig. 250 and based on the molding commands generated in Sec~\ref{Envelope shaping first method kneading commands} and Sec~\ref{Similar gradient method kneading commands}, the kneading command is used to construct the ideal machining PCL mapped onto the end-effector.

The commands  \((h,\, i,\, D_{hi}^{\text{raw}},\, i + N_h'/2,\, D_{h,\,i+N_h'/2}^{\text{raw}})\) are inversely used to generate a point cloud. Then, for each point in the cloud, a semicircular point cloud representing the actual effective area of the end-effector is generated, with each semicircle centered at the corresponding point, perpendicular to XOY plane. Get one point from point cloud, is the circle center \(c_i\), the normal vector \(\mathbf{n_i}\) of the circular surface is the unit vector from the circle center pointing toward the Z-axis, let \(\mathbf{u_i}\) and \(\mathbf{v_i}\) be two unit vectors lying in the plane, perpendicular to each other and perpendicular to the normal vector, because the circle plane perpendicular to XOY plane, so can use the \(\mathbf{u_i}\) is \(\mathbf{(0,0,1)}\), as Equ.\ref{Calculate v by n and u}. Then assume generates \(N_{gp}\) points within the circular area of diameter \(d_m\) end-effector, each generated circular ring is \(r_{gpi}\), as Equ.\ref{Calculate every cirle ring radius}. The number of points on each circular ring varies with the radius (to ensure approximately uniform point density along the ring) as Equ.\ref{Calculate every cirle ring points number}, \(K\) represents the minimum number of points on each circular ring, to avoid shape distortion caused by overly sparse sampling near the center of the ring, \(K\) is typically set to 8, 12, or 16. Points on each ring are evenly distributed in angle as Equ.\ref{Calculate every points angle}. In the end the uniformly distributed point cloud coordinates \(\mathbf{p_{k,j}}\) generated with \(c_i\) as the circle center are show in Equ.\ref{Point cloud generate circle method}.

\begin{equation}
\mathbf{v_i} = \frac{\mathbf{n_i} \times \mathbf{u_i}}{\|\mathbf{n_i} \times \mathbf{u_i}\|}
\label{Calculate v by n and u}
\end{equation}

\begin{equation}
r_{\text{gpi}} =\frac{d_m*k}{2*int(\sqrt{N_{gp}})}, \quad k = 1,2,\dots,\sqrt{N_{gp}}
\label{Calculate every cirle ring radius}
\end{equation}

\begin{equation}
n_{\text{points}, k} = \max\left(K, \, \text{int}\left(\frac{2\pi r_k}{R / (\sqrt{N_{gp}})) }\right)\right)
\label{Calculate every cirle ring points number}
\end{equation}

\begin{equation}
\theta_j = \frac{2\pi j}{n_{\text{points}, k}}, \quad j = 0, 1, \ldots, n_{\text{points}, k} - 1
\label{Calculate every points angle}
\end{equation}

\begin{equation}
\mathbf{p}_{k,j} = \mathbf{c}_i + r_k \cos(\theta_j) \mathbf{u_i} + r_k \sin(\theta_j) \mathbf{v_i}
\label{Point cloud generate circle method}
\end{equation}

\begin{figure}[htbp]
\centering

\begin{minipage}{0.33\textwidth}
\centering
\begin{overpic}[width=\linewidth]{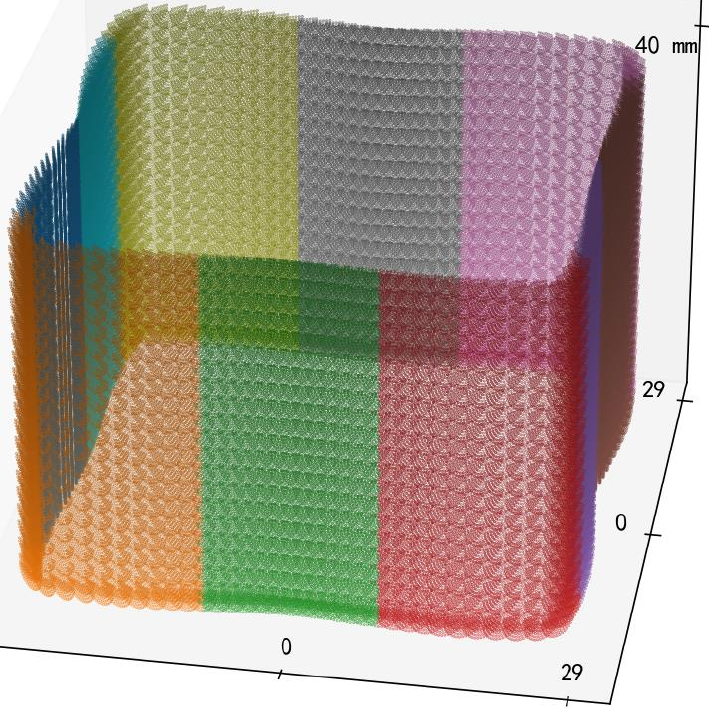}
    \put(3,95){\footnotesize\textbf{A}}
\end{overpic}
\end{minipage}\hfill
\begin{minipage}{0.33\textwidth}
\centering
\begin{overpic}[width=\linewidth]{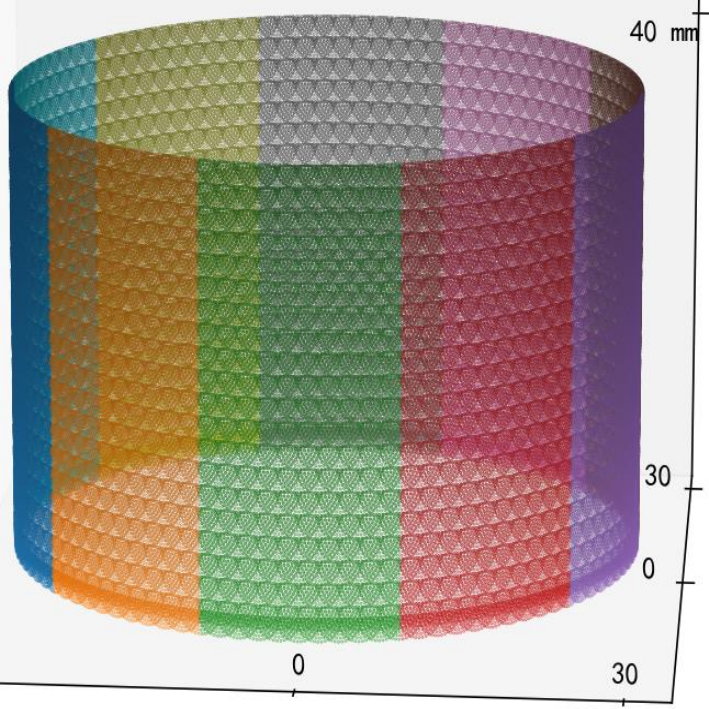}
    \put(3,95){\footnotesize\textbf{B}}
\end{overpic}
\end{minipage}\hfill
\begin{minipage}{0.33\textwidth}
\centering
\begin{overpic}[width=\linewidth]{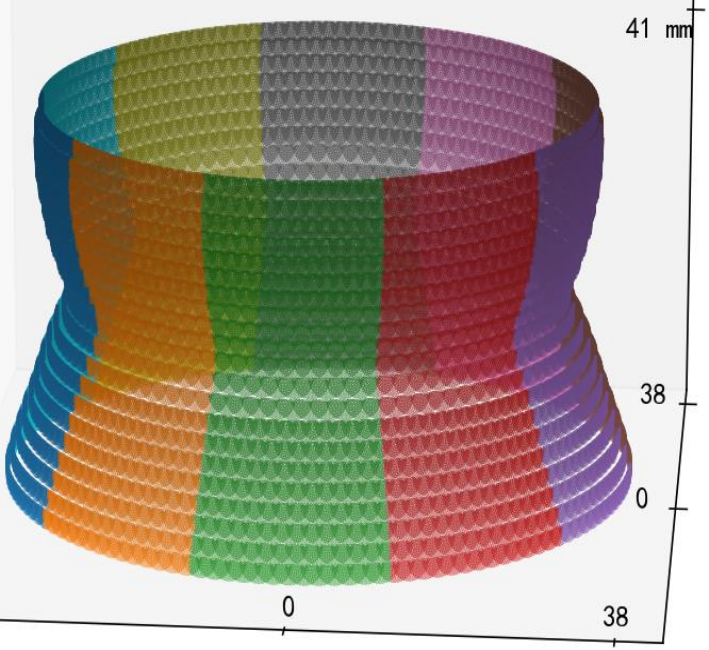}
    \put(3,90){\footnotesize\textbf{C}}
\end{overpic}
\end{minipage}\hfill

\begin{minipage}{0.33\textwidth}
\centering
\begin{overpic}[width=\linewidth]{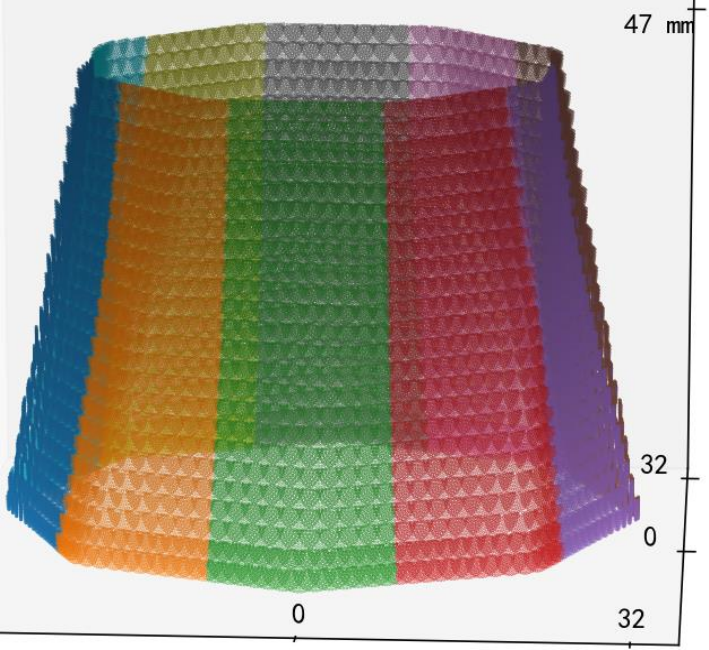}
    \put(0,85){\footnotesize\textbf{D}}
\end{overpic}
\end{minipage}
\begin{minipage}{0.33\textwidth}
\centering
\begin{overpic}[width=\linewidth]{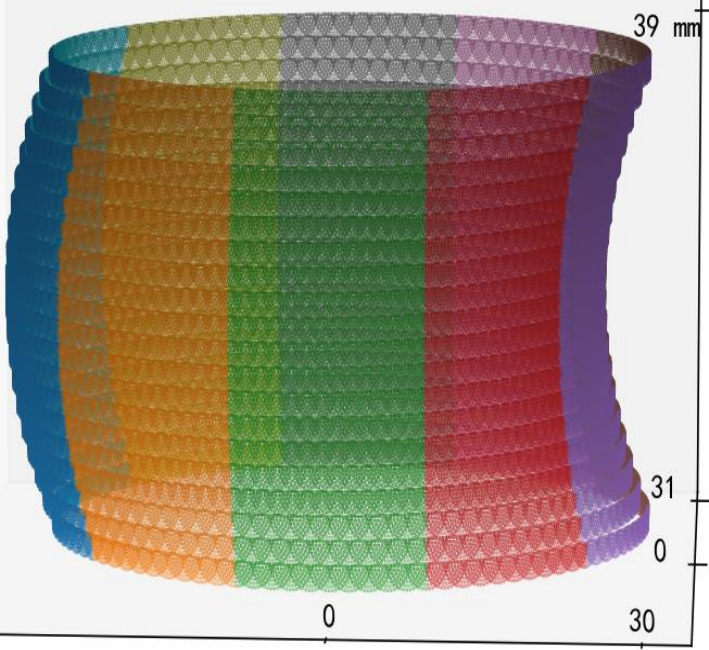}
    \put(0,85){\footnotesize\textbf{E}}
\end{overpic}
\end{minipage}

\caption{Ideal machining point cloud}
\label{Ideal machining point cloud}
\end{figure}

According to Equ.\ref{Point cloud generate circle method}, the generated ideal machining PCL(set \(N_{gp}=40\) ) is shown in Fig~.\ref{Ideal machining point cloud}. A, B, and D are placed in the first layer because they are processed using the enveloping method, C and E are placed in the second layer because they are processed using the similarity method.  

\begin{figure}[H]
\centering
\includegraphics[width=0.75\columnwidth]{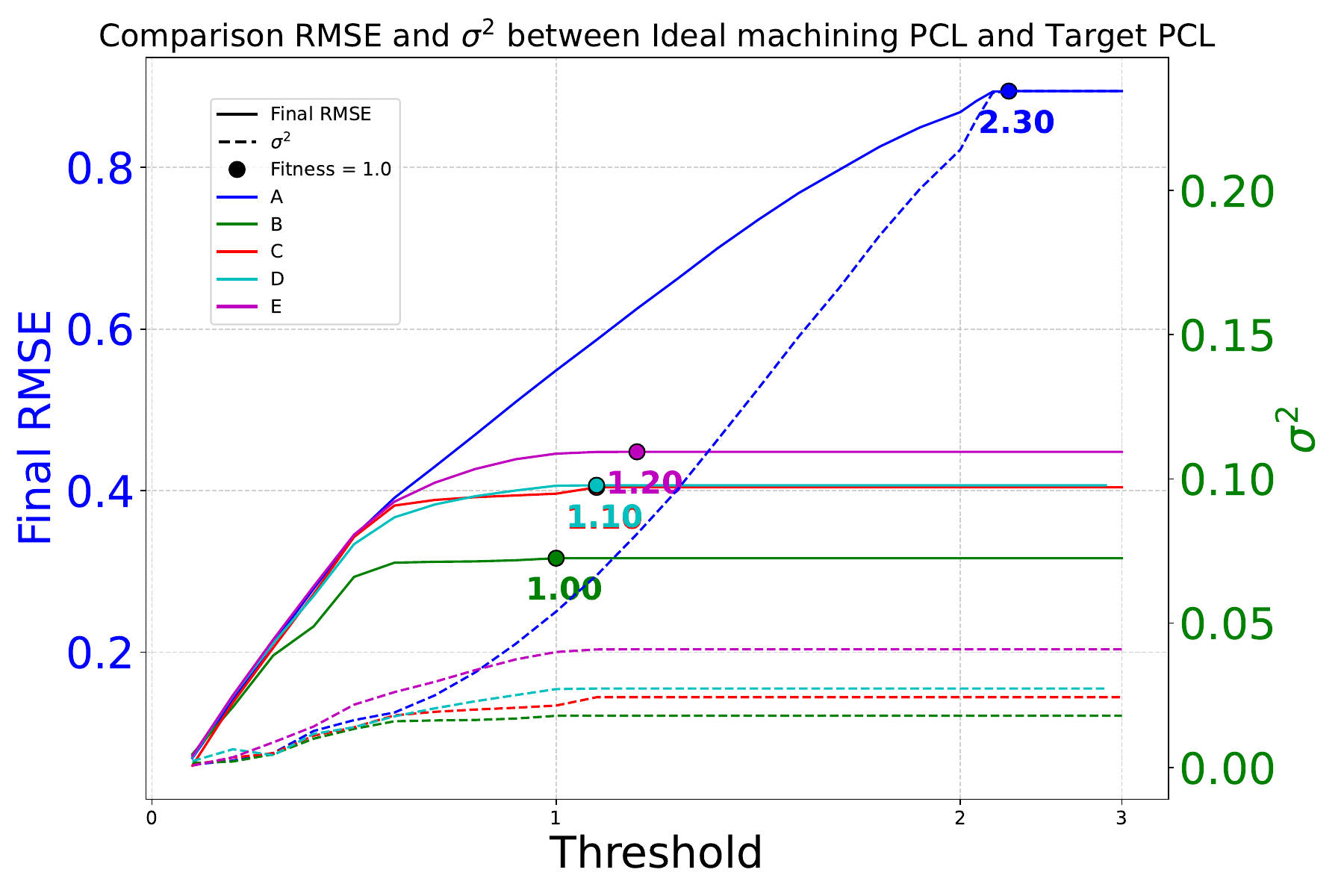}
\caption{\small Comparison RMSE and $\sigma^2$ between Ideal machining PCL and Target PCL}
\label{Ideal machining PCL with target PCL ICP result}
\end{figure}

The ideal machining PCL and the target PCL extracted from the STL model were registered using the ICP algorithm. By adjusting the distance threshold, the fitness was optimized to 1.0. The ideal machining accuracy was evaluated using the obtained root mean square error (RMSE) and residual variance ($\sigma^2$). The variations of ICP registration results with distance threshold for the five geometries A, B, C, D, and E are shown in Fig.~\ref{Ideal machining PCL with target PCL ICP result}. Final registration results of the five geometric objects are shown in the Tab.~\ref{Ideal machining PCL ICP data table}. 

\begin{table}[H]
\centering
\caption{Ideal machining PCL and Target PCL ICP registration results}
\begin{tabular}{ccc}
\hline
Geometric & Threshold & RMSE \\
\hline
A         & 2.3       & 0.895      \\
B         & 1         & 0.316      \\
C         & 1.1       & 0.404      \\
D         & 1.1       & 0.407      \\
E         & 1.2       & 0.448      \\    
\hline
\end{tabular}
\label{Ideal machining PCL ICP data table}
\end{table}

\subsubsection{Actual Kneading result}

\begin{figure}[htbp] 
\centering

\begin{minipage}{0.20\textwidth}
\centering
\begin{overpic}[width=\linewidth]{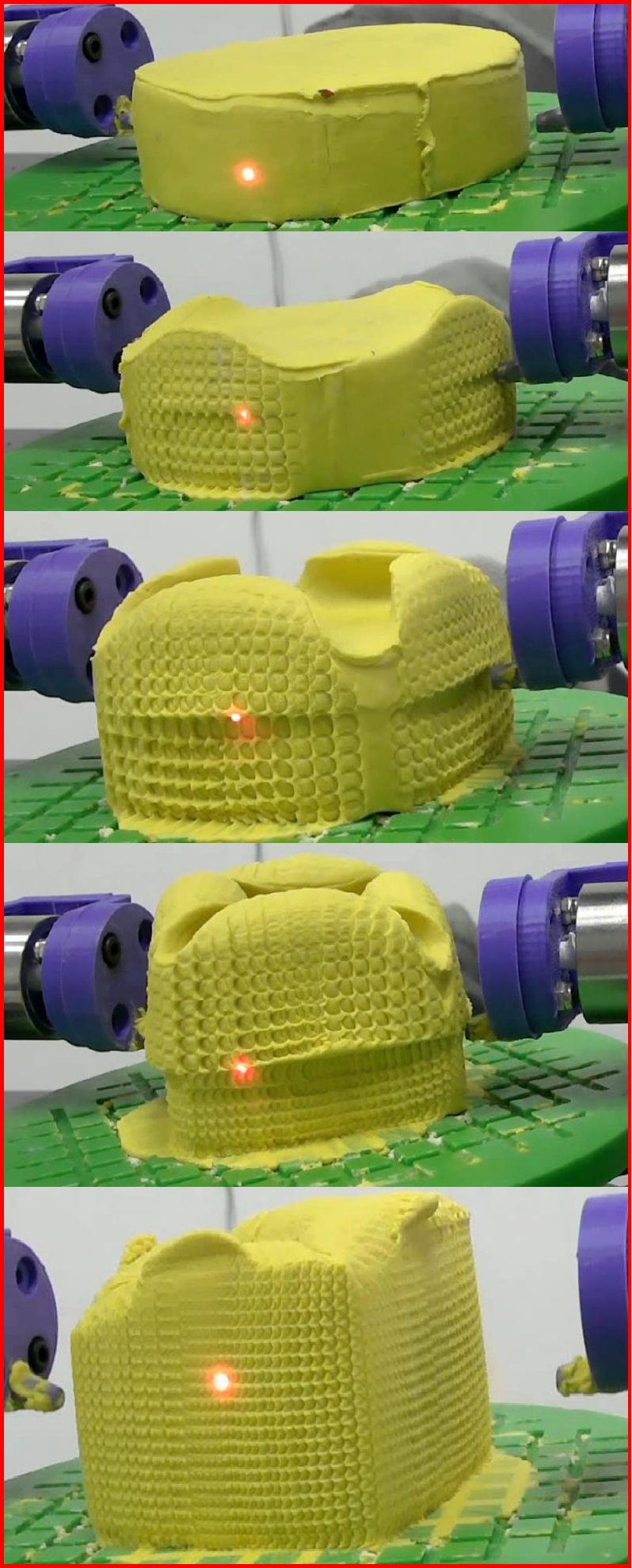}
	\put(-1,96){\scriptsize\colorbox{white}{\textbf{A}}}
\end{overpic}
\end{minipage}\hfill
\begin{minipage}{0.20\textwidth}
\centering
\begin{overpic}[width=\linewidth]{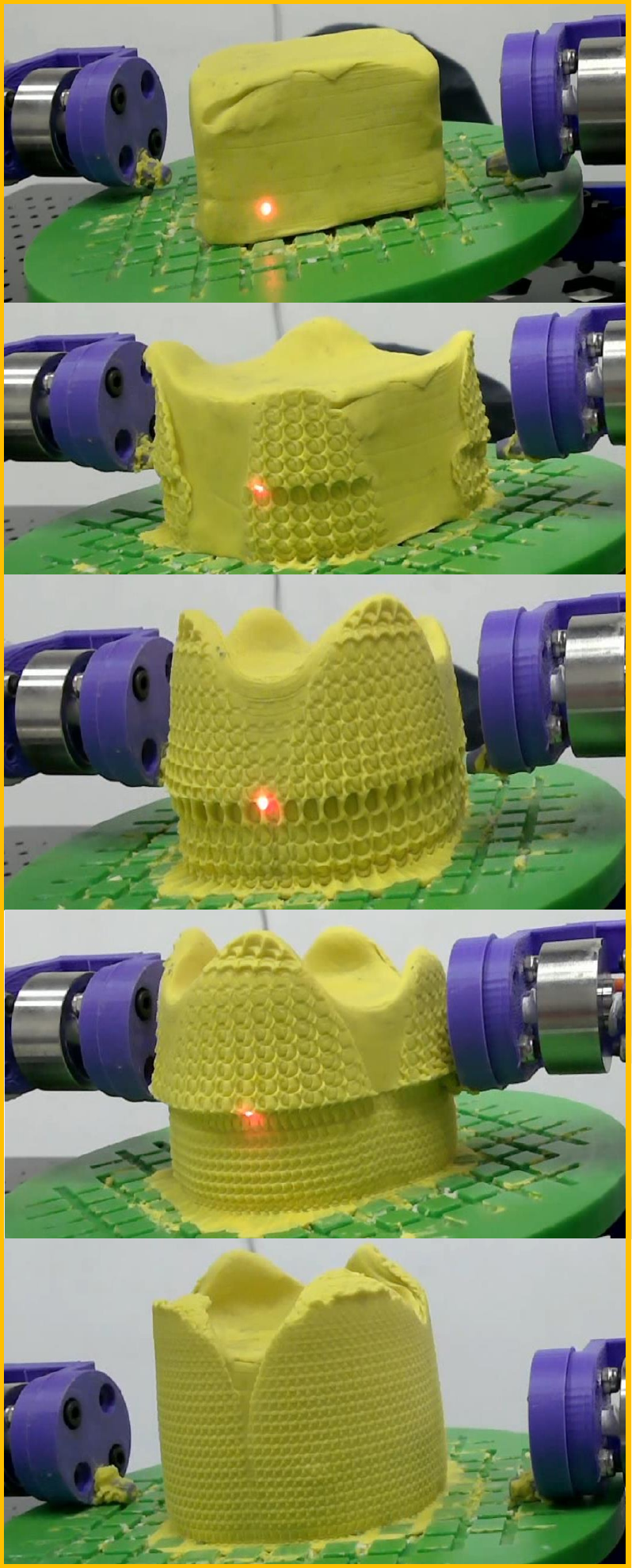}
	\put(0,96){\scriptsize\colorbox{white}{\textbf{B}}}
\end{overpic}
\end{minipage}\hfill
\begin{minipage}{0.20\textwidth}
\centering
\begin{overpic}[width=\linewidth]{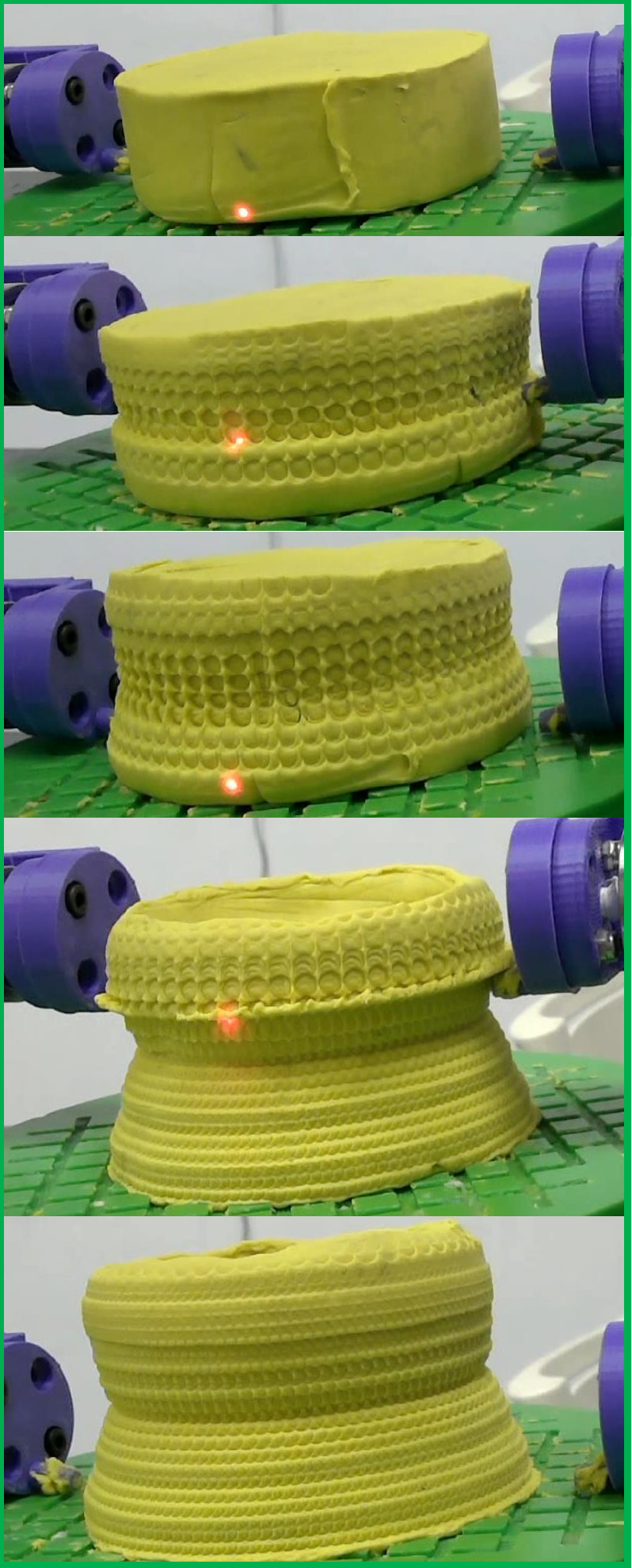}
	\put(0,96){\scriptsize\colorbox{white}{\textbf{C}}}
\end{overpic}
\end{minipage}\hfill
\begin{minipage}{0.20\textwidth}
\centering
\begin{overpic}[width=\linewidth]{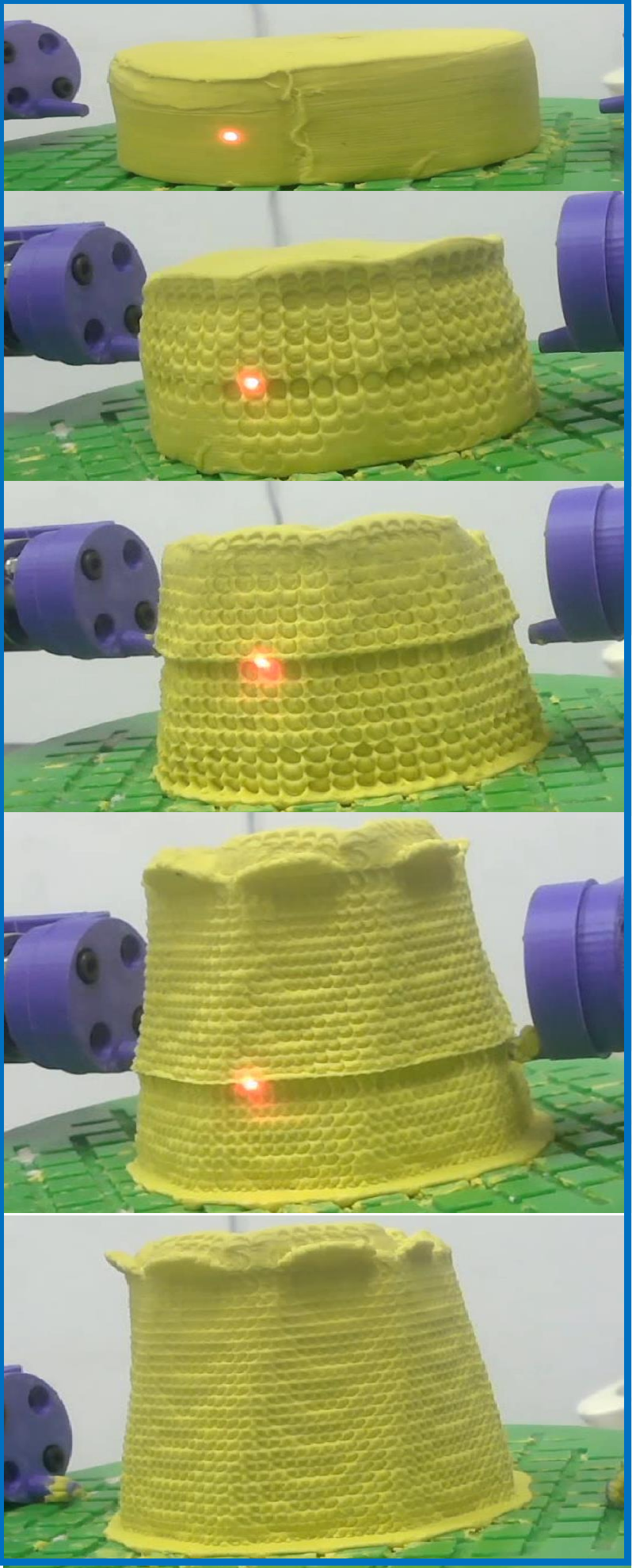}
	\put(0,96){\scriptsize\colorbox{white}{\textbf{D}}}
\end{overpic}
\end{minipage}\hfill
\begin{minipage}{0.20\textwidth}
\centering
\begin{overpic}[width=\linewidth]{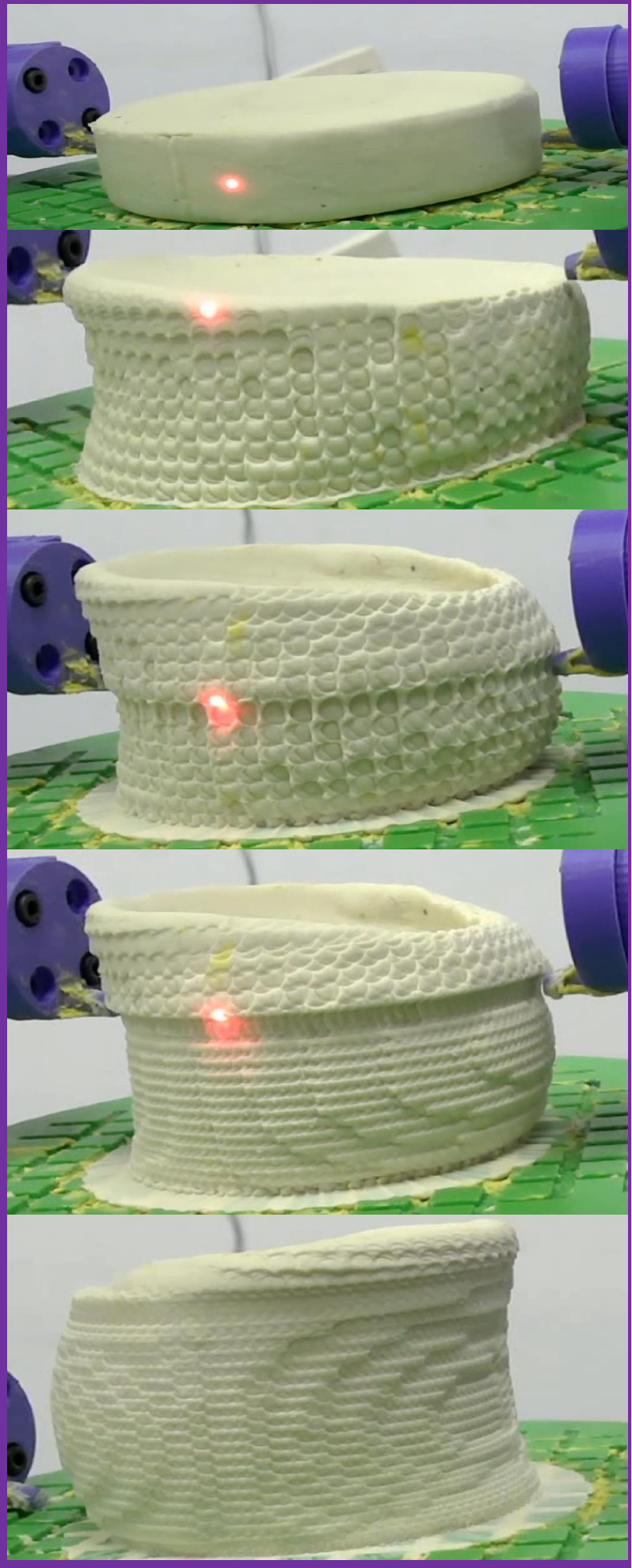}
	\put(0,96){\scriptsize\colorbox{white}{\textbf{E}}}
\end{overpic}
\end{minipage}

\caption{Geometric actual knead result}
\label{Geometric actual knead result}

\end{figure}

To evaluate the geometric fidelity of the kneading-based forming process, the actual kneading point clouds (Actual Kneading PCLs) of Geometries A--E are systematically compared with the corresponding 3D-printed reference point clouds (Target PCLs), as well as with the compensated point clouds obtained through RMSE-based correction. The kneading-based forming processes of Geometries A, B, C, D, and E are illustrated in Fig.~\ref{Geometric actual knead result}. Geometry A demonstrates the transformation of a circular blank with an initial diameter of 80~mm into a square prism, while Fig.~\ref{Geometric actual knead result}-B shows the forming process from a square-based blank with a diagonal length of 84.5~mm into a cylindrical geometry. Fig.~\ref{Geometric actual knead result}-C, D, and E present the kneading processes of the corresponding geometries using the Similar Gradient method. These results visually confirm that the proposed kneading platform can realize both envelope-type and non-envelope-type geometries through adaptive kneading strategies.

\begin{figure}[htbp] 
\centering

\begin{minipage}{0.20\textwidth}
\centering
\begin{overpic}[width=\linewidth]{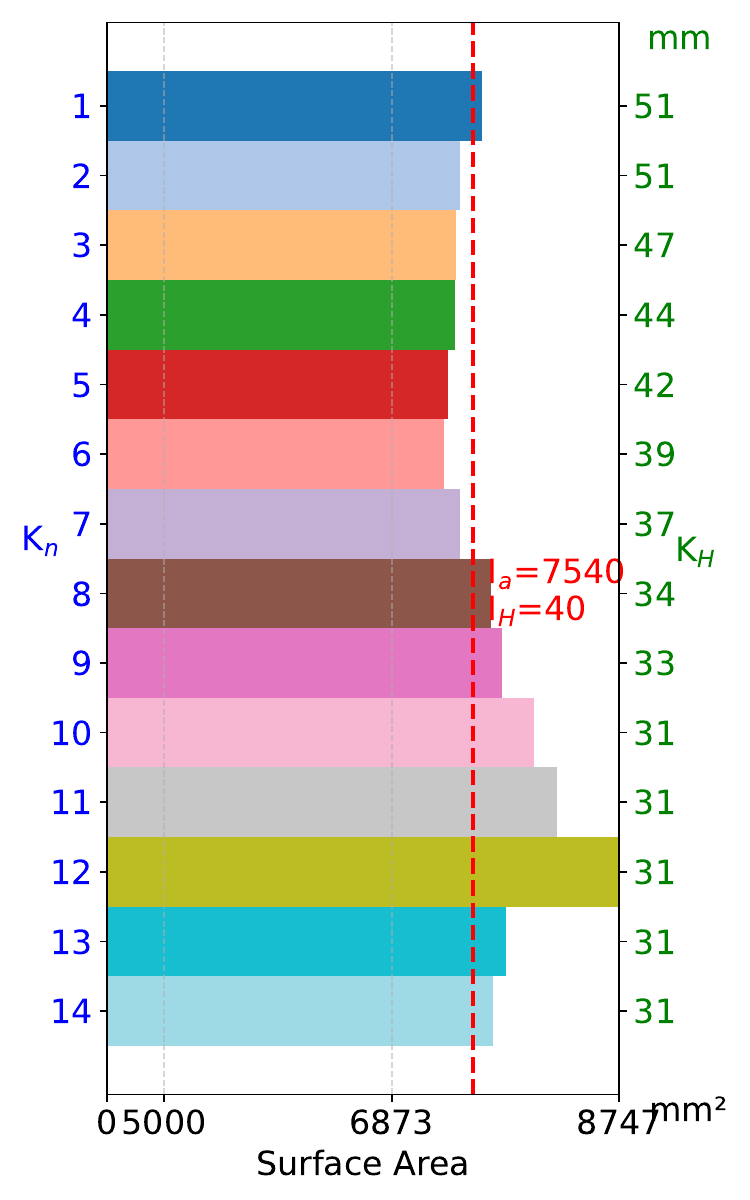}
	\put(0,99){\scriptsize\textbf{A}}
\end{overpic}
\end{minipage}\hfill
\begin{minipage}{0.20\textwidth}
\centering
\begin{overpic}[width=\linewidth]{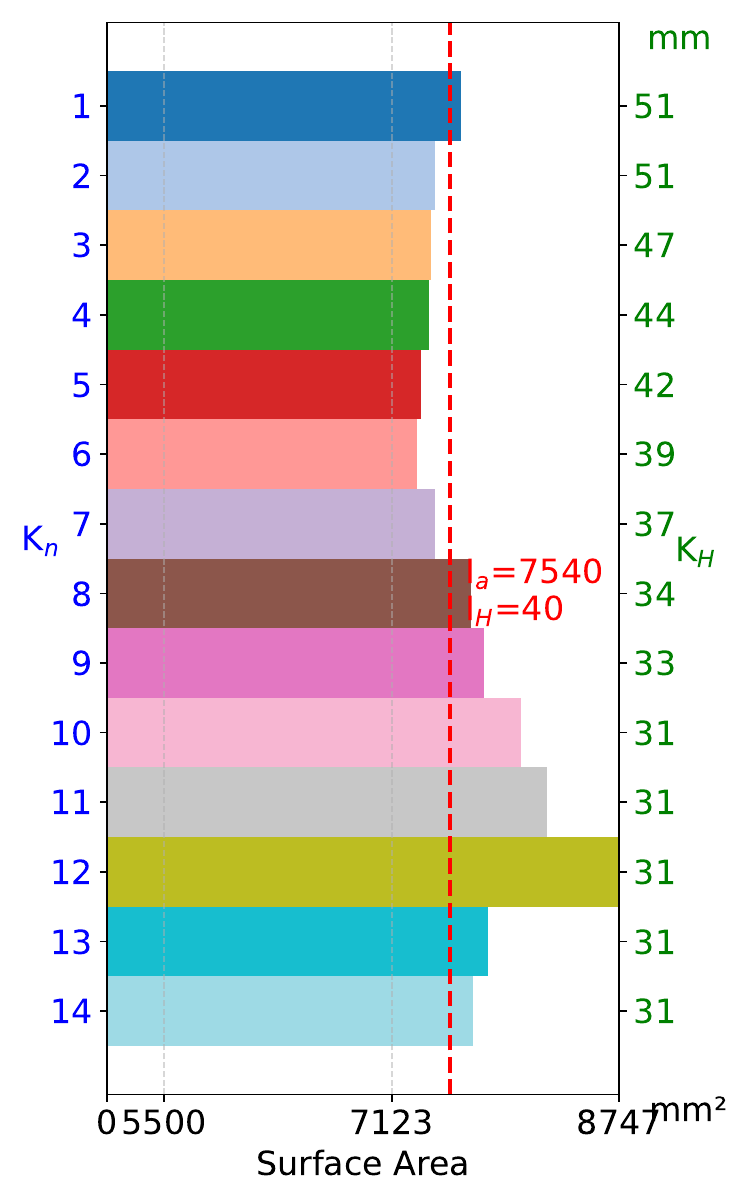}
	\put(0,99){\scriptsize\textbf{B}}
\end{overpic}
\end{minipage}\hfill
\begin{minipage}{0.20\textwidth}
\centering
\begin{overpic}[width=\linewidth]{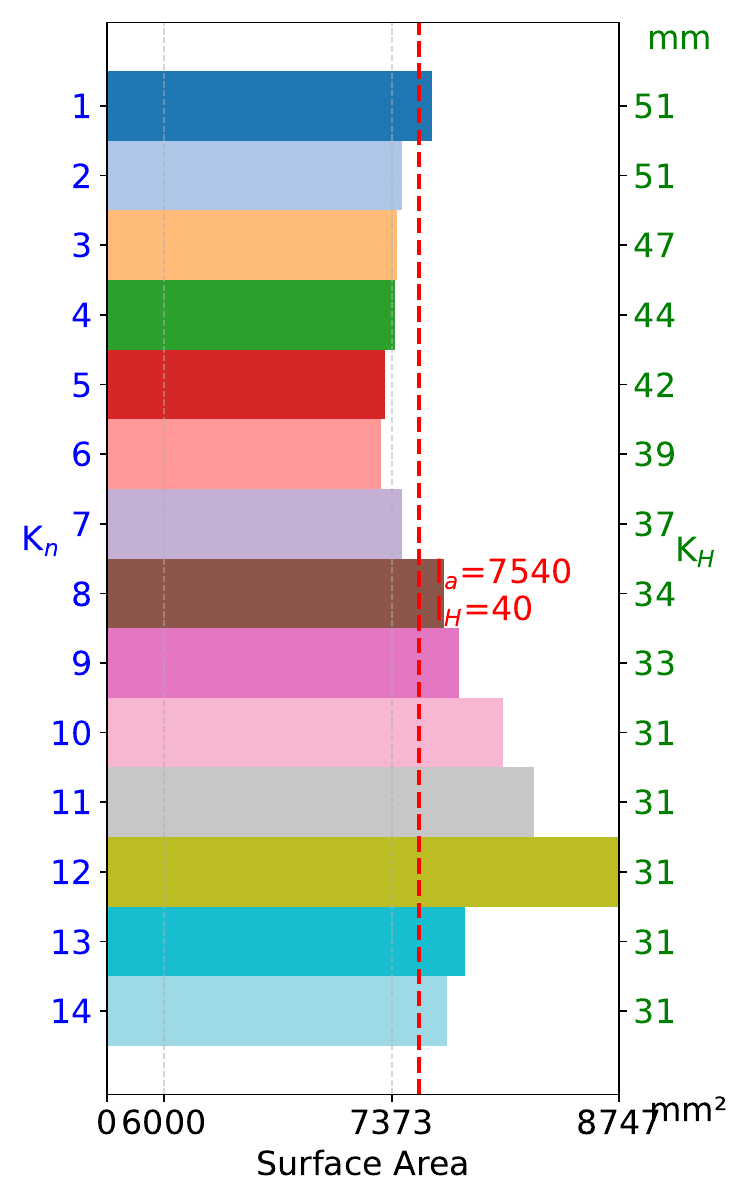}
	\put(0,99){\scriptsize\textbf{C}}
\end{overpic}
\end{minipage}\hfill
\begin{minipage}{0.20\textwidth}
\centering
\begin{overpic}[width=\linewidth]{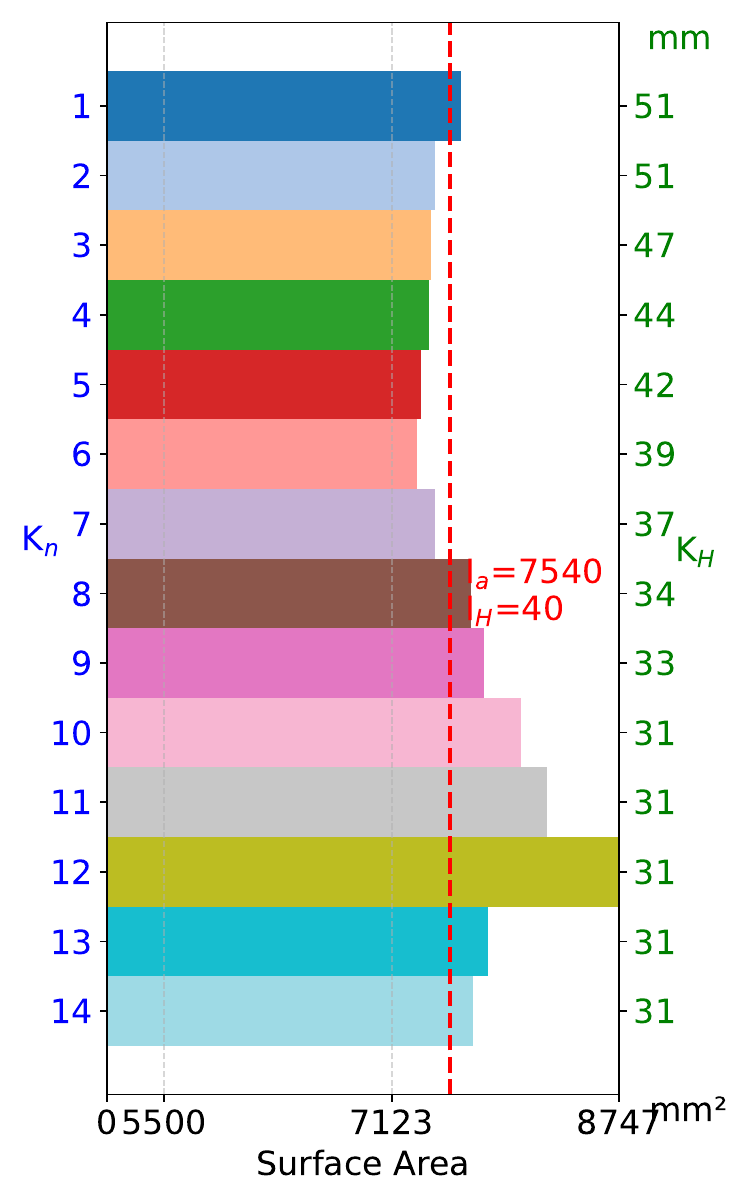}
	\put(0,99){\scriptsize\textbf{D}}
\end{overpic}
\end{minipage}\hfill
\begin{minipage}{0.20\textwidth}
\centering
\begin{overpic}[width=\linewidth]{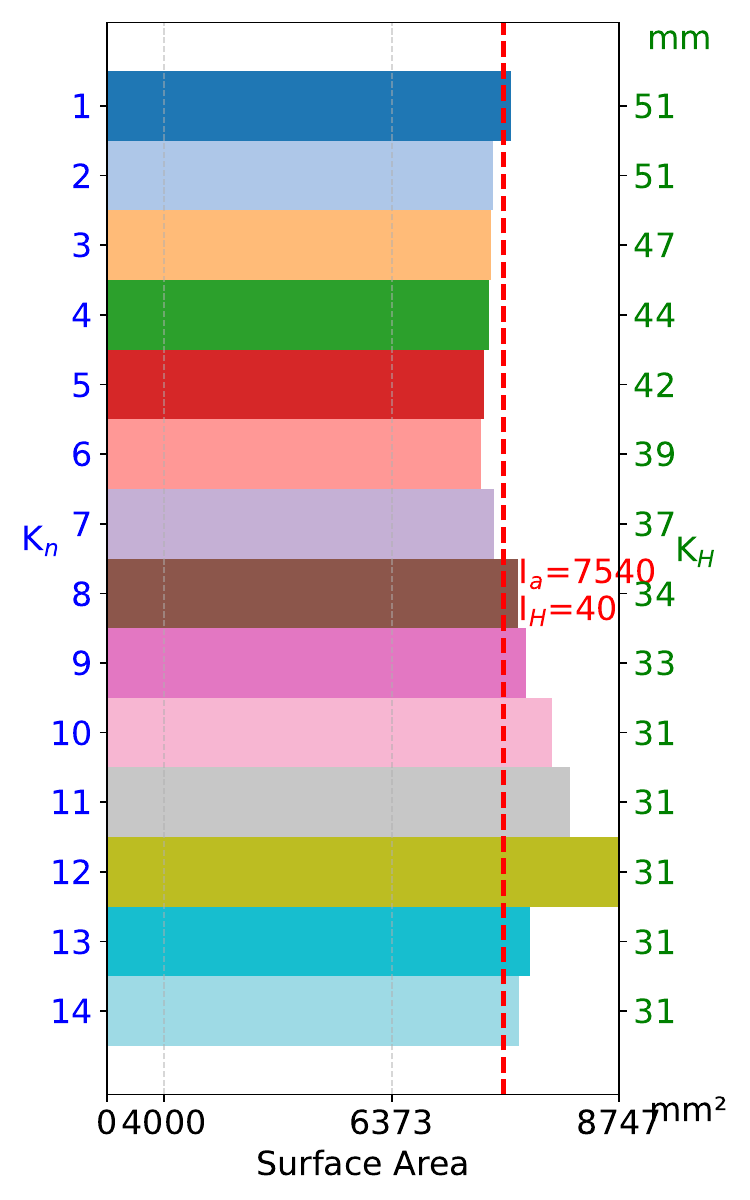}
	\put(0,99){\scriptsize\textbf{E}}
\end{overpic}
\end{minipage}

\caption{Geometric Knead area of surface}
\label{Geometric Knead area of surface}

\end{figure}

\begin{figure}[htbp]
\centering

\begin{minipage}{0.23\textwidth}
\centering
\begin{overpic}[width=\linewidth]{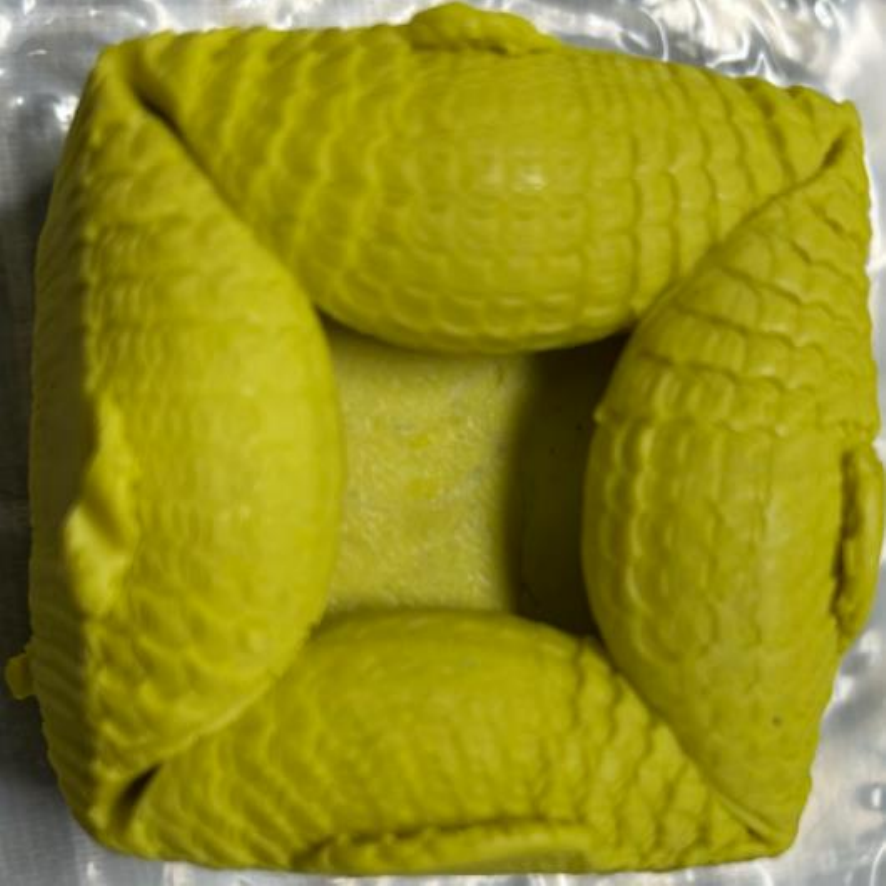}
    \put(0,91){\scriptsize\colorbox{white}{\textbf{$K_{A}$}}}
\end{overpic}
\end{minipage}
\hspace{0.01\textwidth}
\begin{minipage}{0.23\textwidth}
\centering
\begin{overpic}[width=\linewidth]{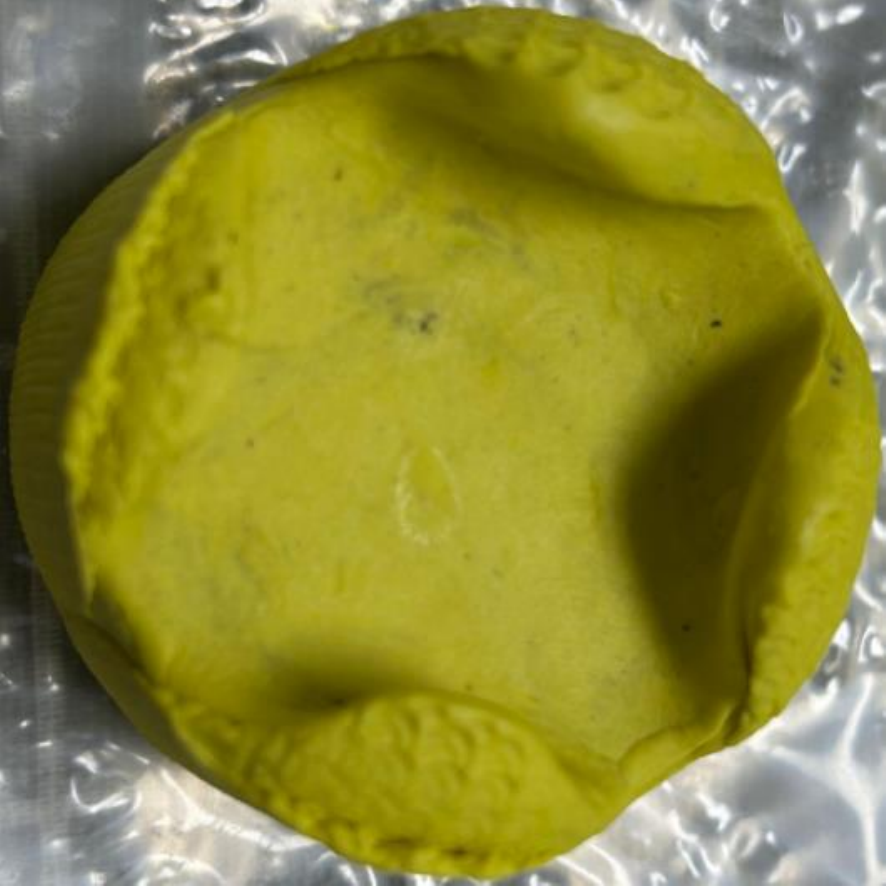}
    \put(-1,91){\scriptsize\colorbox{white}{\textbf{$K_{B}$}}}
\end{overpic}
\end{minipage}
\hspace{0.01\textwidth}
\begin{minipage}{0.23\textwidth}
\centering
\begin{overpic}[width=\linewidth]{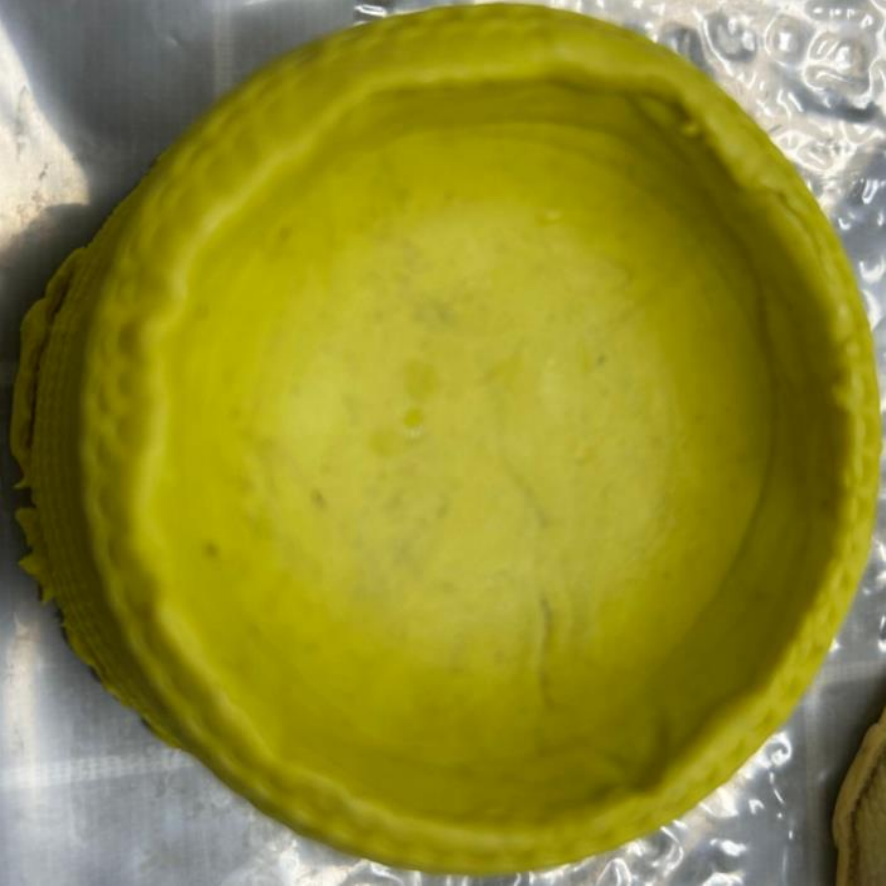}
    \put(0,91){\scriptsize\colorbox{white}{\textbf{$K_{C1}$}}}
\end{overpic}
\end{minipage}
\begin{minipage}{0.23\textwidth}
\centering
\begin{overpic}[width=\linewidth]{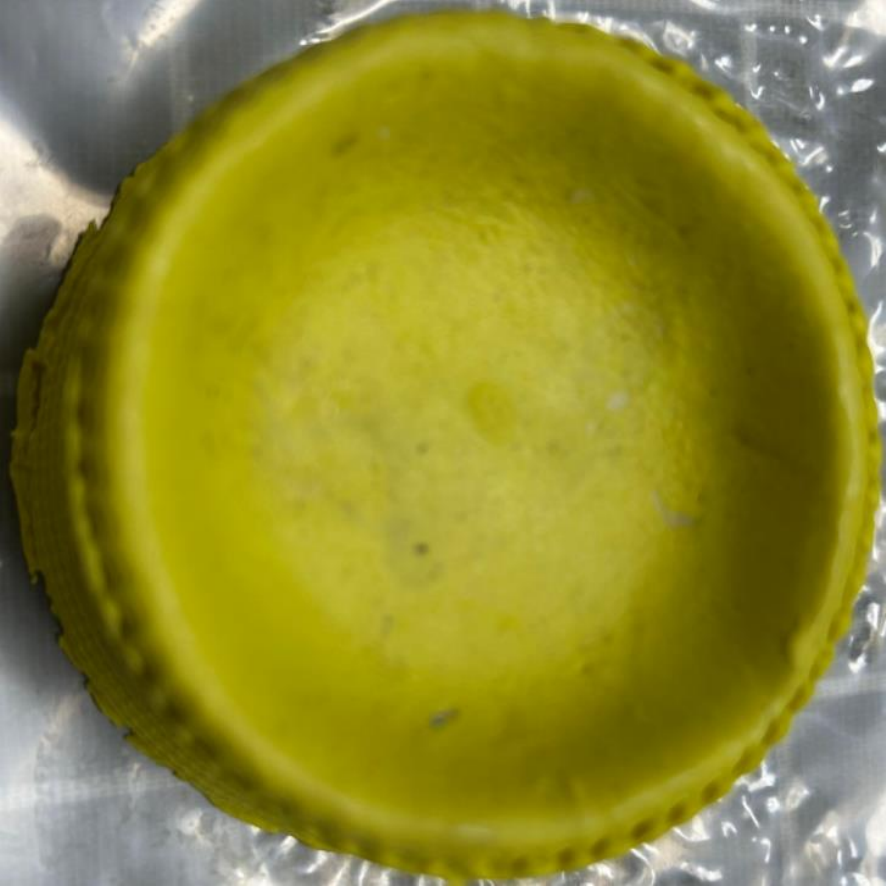}
    \put(-1,91){\scriptsize\colorbox{white}{\textbf{$K_{C2}$}}}
\end{overpic}
\end{minipage}

\begin{minipage}{0.23\textwidth}
\centering
\begin{overpic}[width=\linewidth]{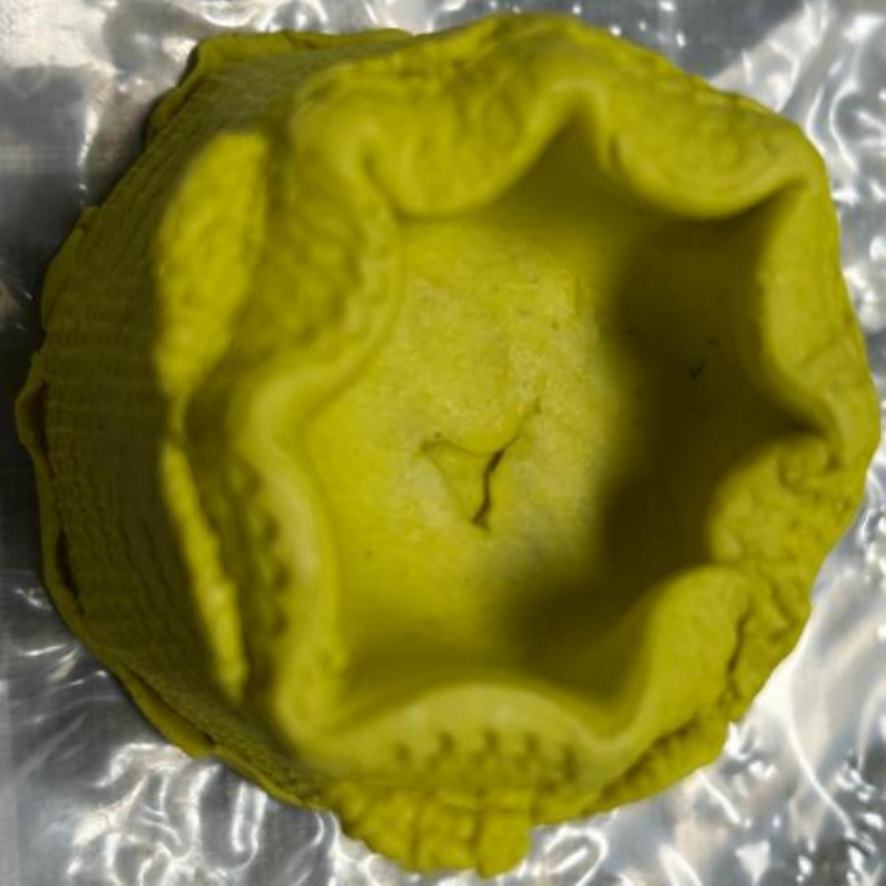}
    \put(-1,91){\scriptsize\colorbox{white}{\textbf{$K_{D1}$}}}
\end{overpic}
\end{minipage}
\begin{minipage}{0.23\textwidth}
\centering
\begin{overpic}[width=\linewidth]{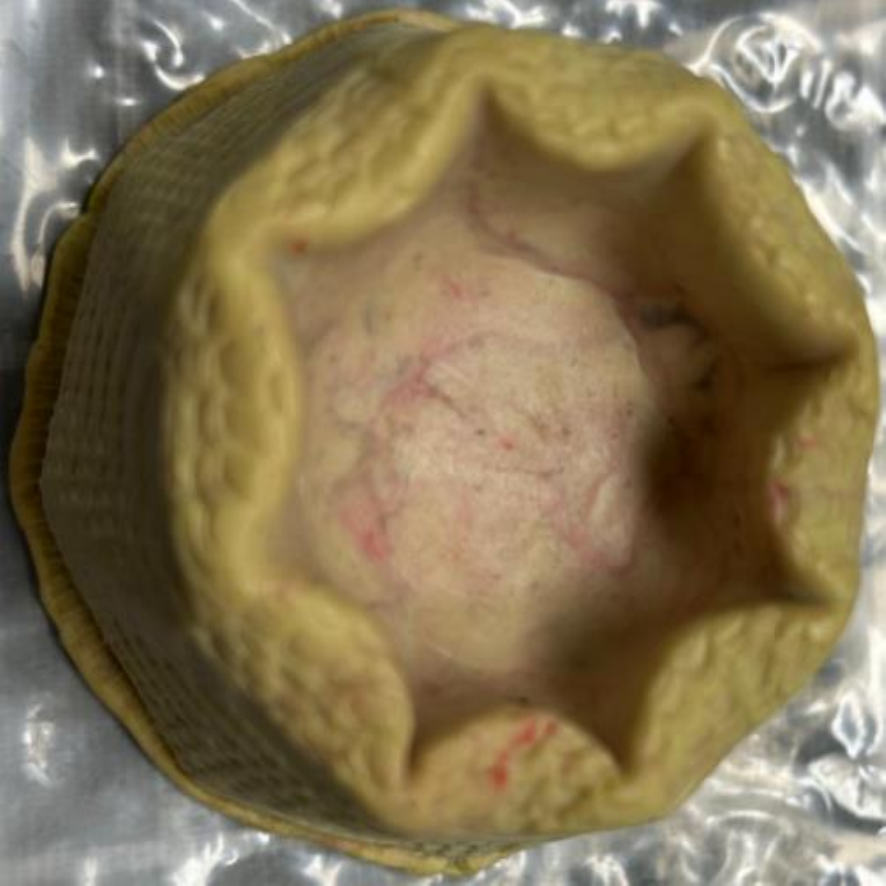}
    \put(0,91){\scriptsize\colorbox{white}{\textbf{$K_{D2}$}}}
\end{overpic}
\end{minipage}
\hspace{0.01\textwidth}
\begin{minipage}{0.23\textwidth}
\centering
\begin{overpic}[width=\linewidth]{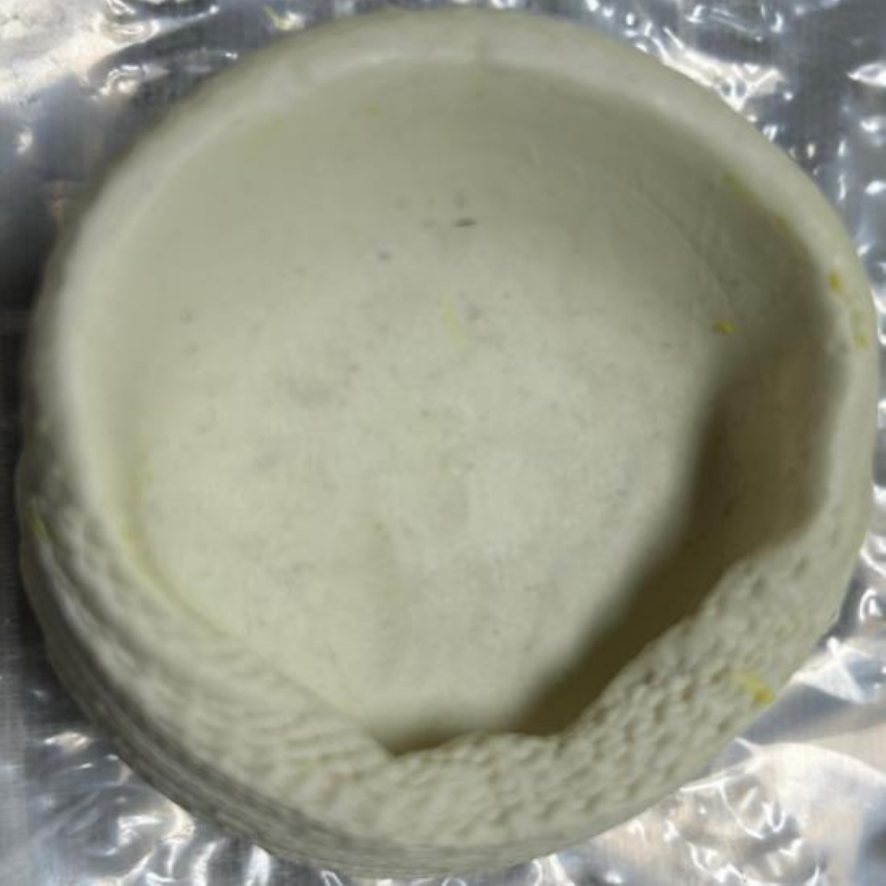}
    \put(0,91){\scriptsize\colorbox{white}{\textbf{$K_{E1}$}}}
\end{overpic}
\end{minipage}
\begin{minipage}{0.23\textwidth}
\centering
\begin{overpic}[width=\linewidth]{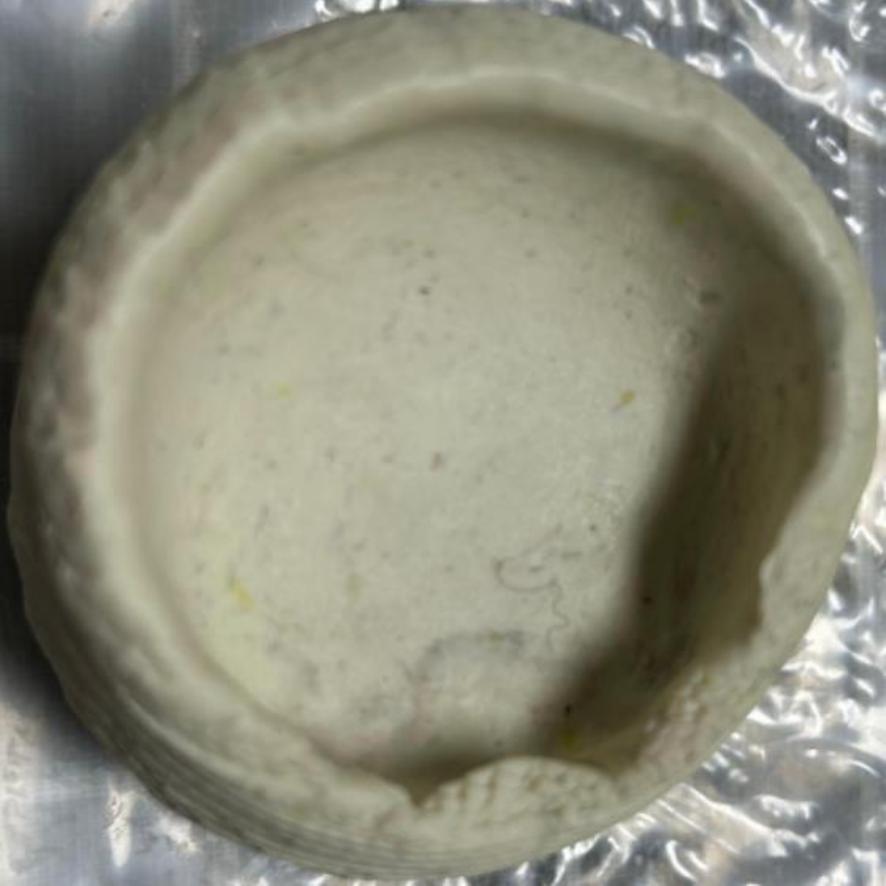}
    \put(0,91){\scriptsize\colorbox{white}{\textbf{$K_{E2}$}}}
\end{overpic}
\end{minipage}

\caption{Geometric model top is hollowed}
\label{Geometric model top is hollowed}
\end{figure}

During the forming process, the evolution of the blank surface area is shown in Fig.~\ref{Geometric Knead area of surface}. For all five geometries, the final formed blanks exhibit surface areas and heights larger than the corresponding ideal values. This deviation is mainly attributed to two factors, as illustrated in Fig.~\ref{Geometric model top is hollowed}: (i) material flow on the blank surface during kneading, which induces cavities in the upper regions of the formed geometries, and (ii) elastic rebound of the material after the completion of kneading. These effects jointly introduce systematic geometric deviations in the actual kneading results.

\begin{figure}[H]
\centering
\includegraphics[width=0.8\columnwidth]{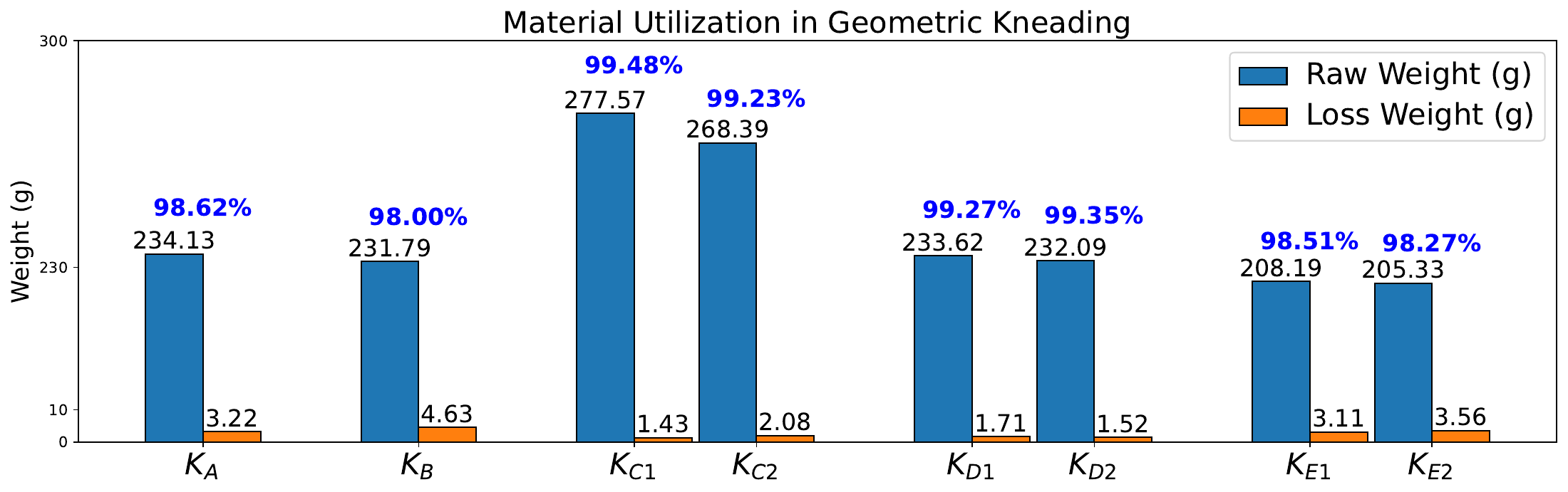}
\caption{\small Geometric material mass change before and after kneading
}
\label{Geometric material mass change before and after kneading}
\end{figure}

\begin{figure}[H]
\centering
\includegraphics[width=0.8\columnwidth]{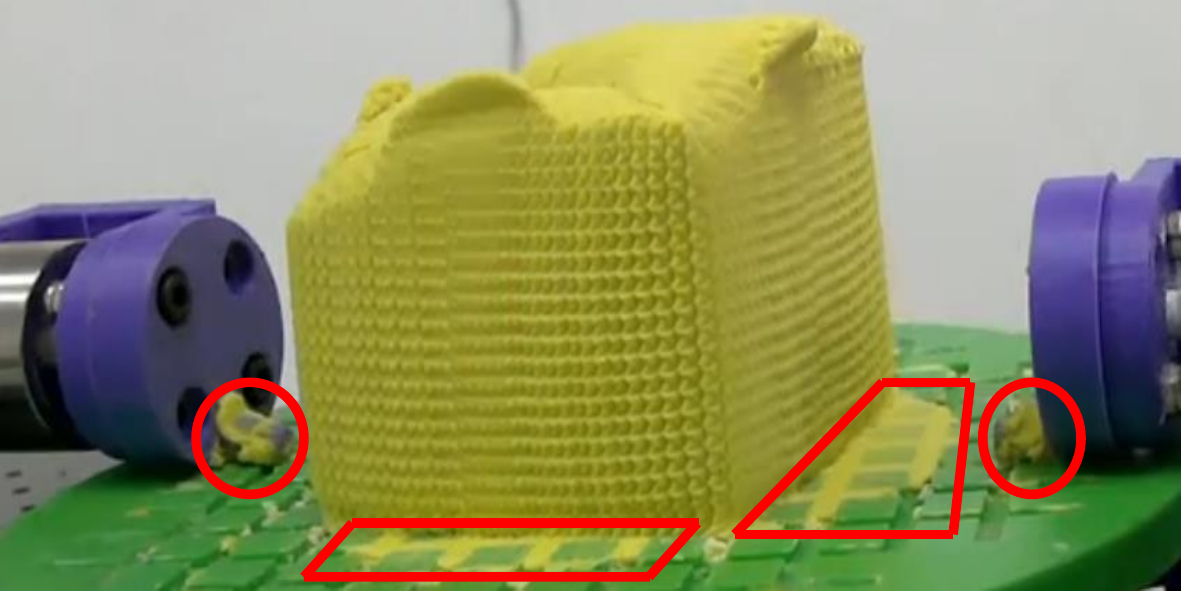}
\caption{\small Material loss in the kneading process}
\label{Material loss in the kneading process}
\end{figure}

Material utilization during the kneading process is quantified by comparing the material mass before and after forming, as shown in Fig.~\ref{Geometric material mass change before and after kneading}. For Geometries A and B, as well as Geometries C, D, and E formed using two different kneading strategies, the material utilization consistently exceeds 98\%. The primary sources of material loss are illustrated in Fig.~\ref{Material loss in the kneading process}, including adhesion between the fingers and the material and material detachment caused by the grooved structure of the rotating frustum platform. A comparative summary of material utilization for conventional subtractive manufacturing, additive manufacturing, and the proposed kneading-based forming method is provided in Tab.~\ref{Material utilization different methods}, highlighting the material efficiency advantage of the proposed approach.

\begin{table}[H]
\centering
\caption{Material utilization efficiency of different manufacturing methods}
\label{Material utilization different methods}
\begin{tabular}{lcccc}
\hline
Process & Utilization (\%) & Scrap Type & Recyclability \\
\hline
SM & $\sim$30--60 & Chips & Medium \\
AM & $\sim$70--90 & Support waste & Low--Medium \\
\textbf{Our} & \textbf{$\ge$98} & None & High \\
\hline
\end{tabular}
\end{table}

\begin{figure}[H]
\centering
\includegraphics[width=0.7\columnwidth]{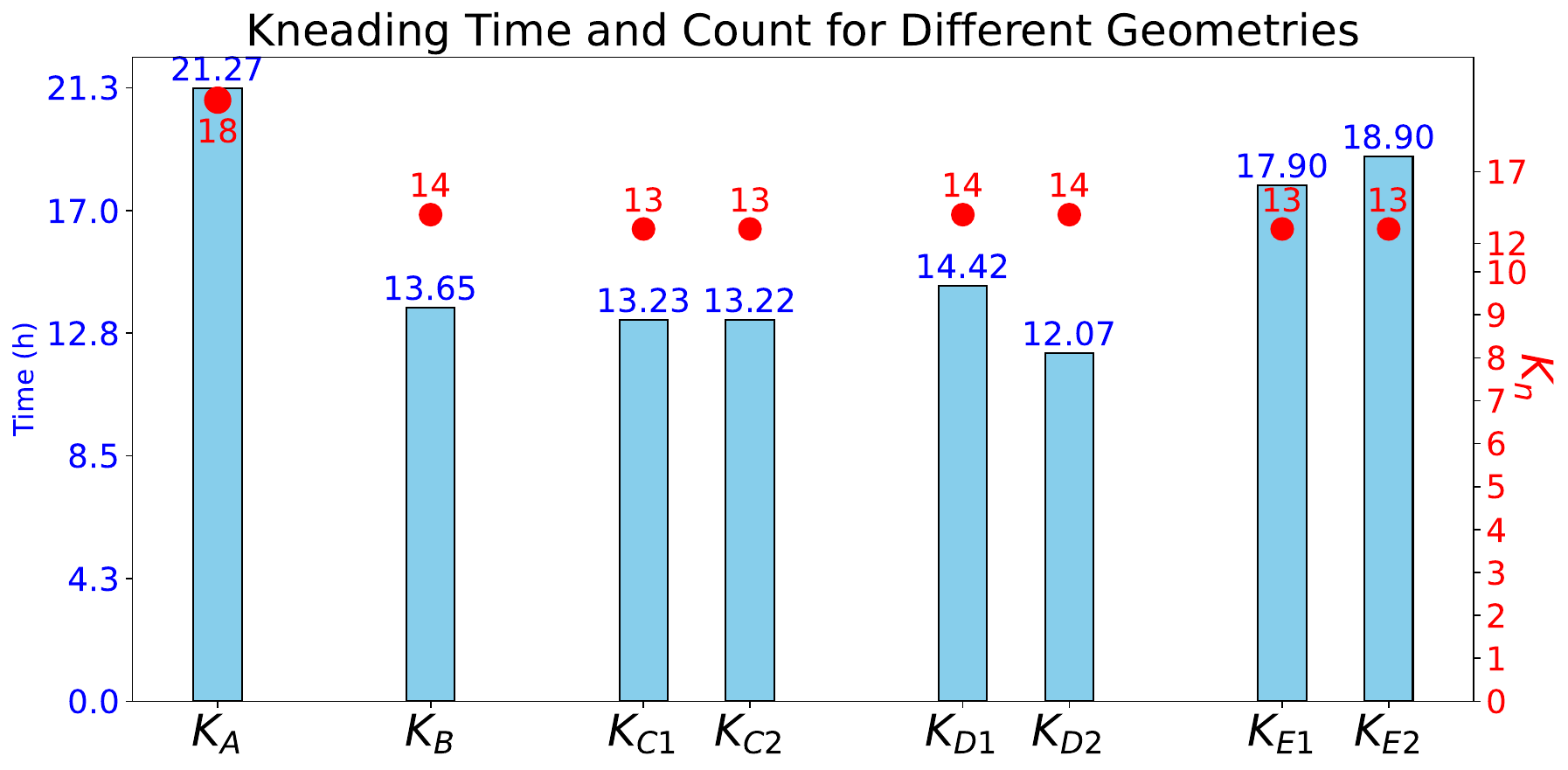}
\caption{\small Geometric knead time and knead cycles}
\label{Geometric knead time and knead cycles}
\end{figure}

The processing time and the number of kneading cycles for each geometry are presented in Fig.~\ref{Geometric knead time and knead cycles}. The required number of processing cycles is primarily determined by the dimensional difference between the initial blank and the target geometry. Geometry A exhibits the largest dimensional difference and therefore requires the greatest number of kneading cycles, resulting in the longest processing time. For Geometry C, the kneading times of the two methods are nearly identical. In contrast, for Geometry D, the Similar Gradient method requires approximately one additional hour compared to the Envelope Shaping First method, whereas for Geometry E, the Envelope Shaping First method requires approximately two additional hours. These results indicate that, under identical initial blank dimensions and processing cycle counts, the intrinsic geometric characteristics of the object dominate the overall processing time.

As shown in Fig.~\ref{Ideal machining PCL with target PCL ICP result} and Tab.~\ref{Ideal machining PCL ICP data table}, the Threshold values at which the fitness reaches 1.0 are close for Geometries B, C, D, and E, while Geometry A exhibits the largest Threshold at the same fitness level. This difference is primarily related to cross-sectional geometry: B, C, and E have circular cross-sections, D has an octagonal cross-section approximating a circle, whereas A has a square cross-section that deviates most from circularity, leading to reduced ideal machining accuracy.

After each kneading operation with the preset cutting depth from bottom to top, the entire blank is scanned. The actual machining accuracy is evaluated by performing ICP registration between the point cloud obtained from the final molded blank scan and the target PCL. The machining accuracy is quantified using the RMSE and $\sigma^2$ obtained from the ICP registration. For consistency, the actual kneading PCLs used for comparison are extracted at the same height as the corresponding target PCLs.

For Geometries A and B, the kneading process involves a discrete transformation of the cross-sectional shape between the initial blank and the target geometry. Specifically, Geometry A is formed from a cylindrical blank with a circular cross-section into a square prism, whereas Geometry B is formed from a square-based blank into a cylindrical geometry. Such non-continuous geometric transitions introduce additional deformation complexity compared to geometries with continuous cross-sectional evolution.

\begin{figure}[H]
\centering
\includegraphics[width=0.75\columnwidth]{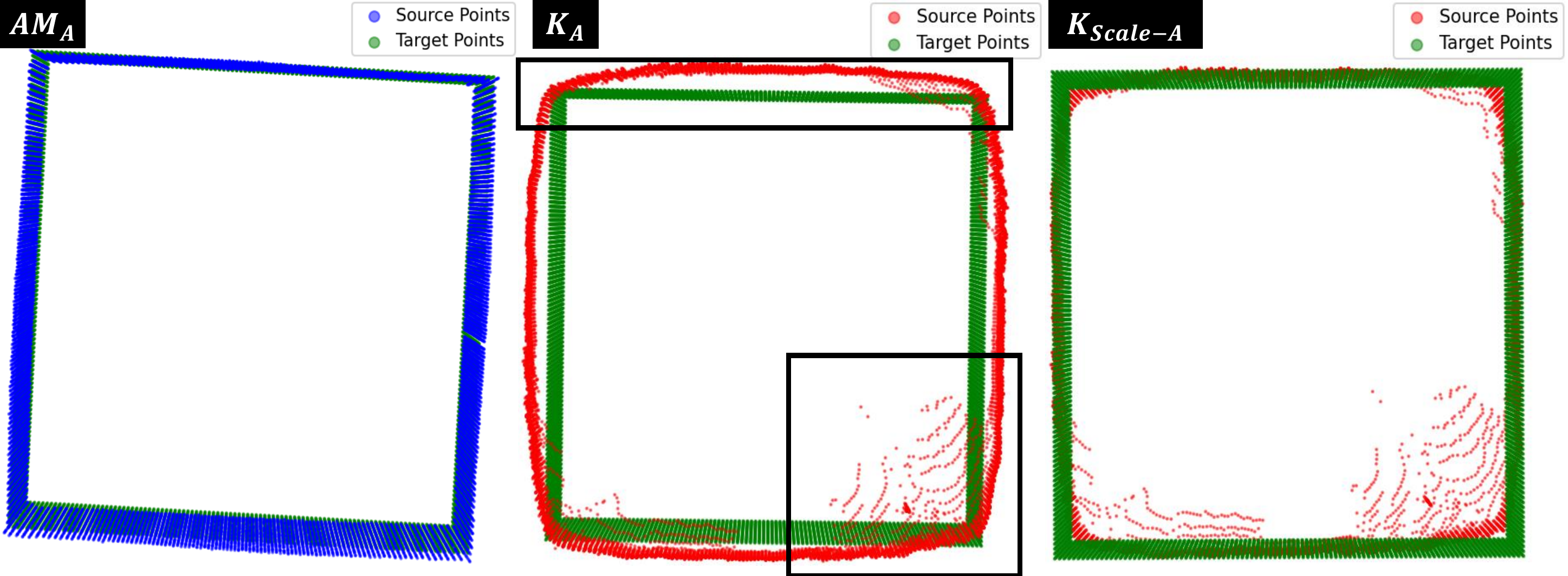}
\caption{\small Geometric A: Registration results between Actual knead, Compensate PCL and Target PCL}
\label{Geometric A actual and scale PCL}
\end{figure}

As shown in Fig.~\ref{Geometric A actual and scale PCL}, Geometry A exhibits a pronounced global scale expansion after kneading. The deviation between the actual kneaded PCL and the target PCL is spatially uniform and extends across both the sidewalls and top region. This behavior indicates that the dominant error source in Geometry A is elastic rebound of the material after the completion of kneading. Moreover, due to the transformation from a circular to a square cross-section, stress redistribution is amplified in the corner regions, leading to larger RMSE and more irregular variations in $\sigma^2$ compared with other geometries. After applying RMSE-based compensation, the compensated PCL shows a significantly improved overlap with the target PCL.

\begin{figure}[htbp]
\centering

\begin{minipage}{0.75\textwidth}
\centering
\begin{overpic}[width=\linewidth]{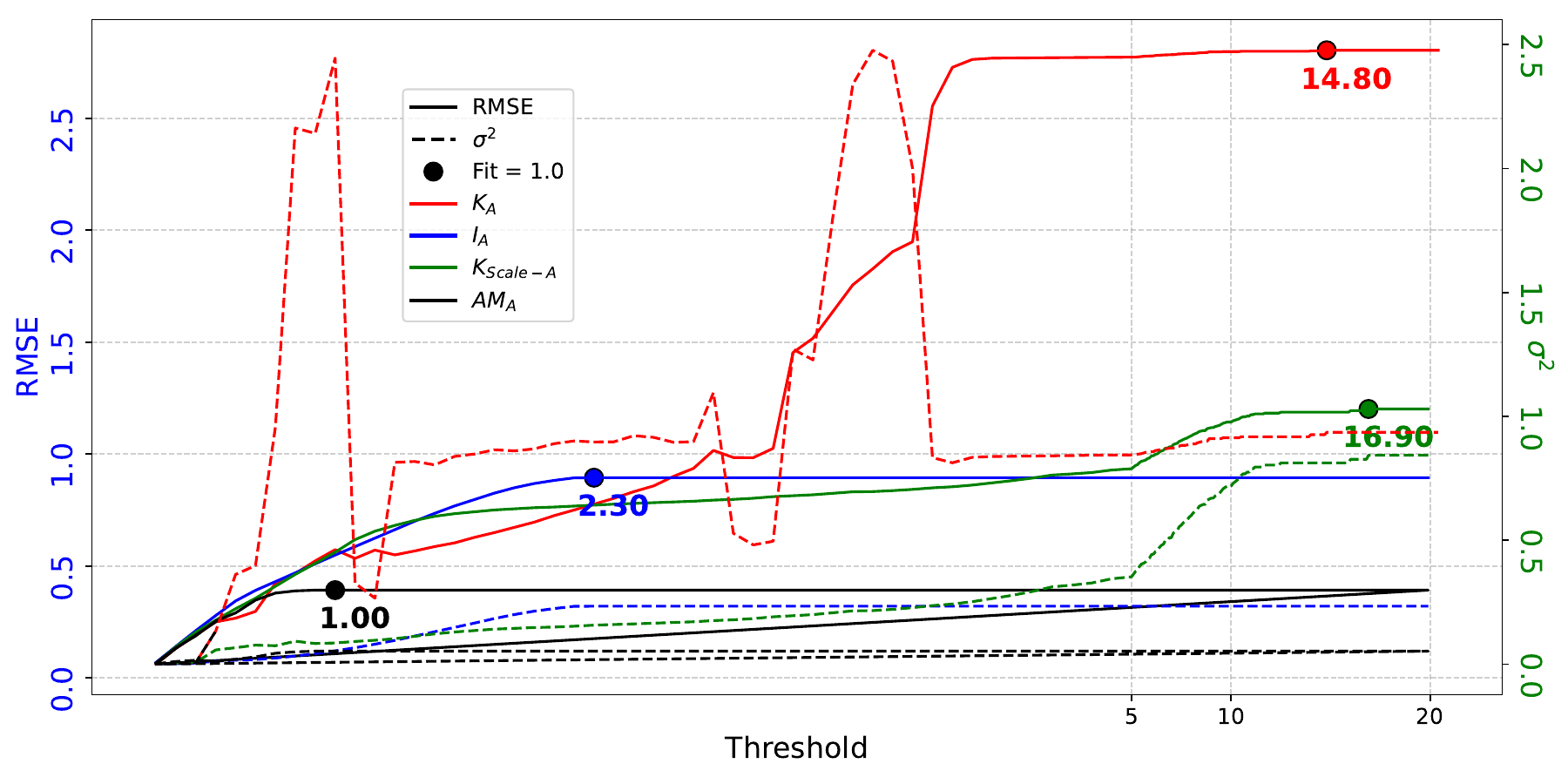}
    \put(0,50){\footnotesize\textbf{I}}
\end{overpic}
\end{minipage}\hfill
\begin{minipage}{0.25\textwidth}
\centering
\begin{overpic}[width=\linewidth]{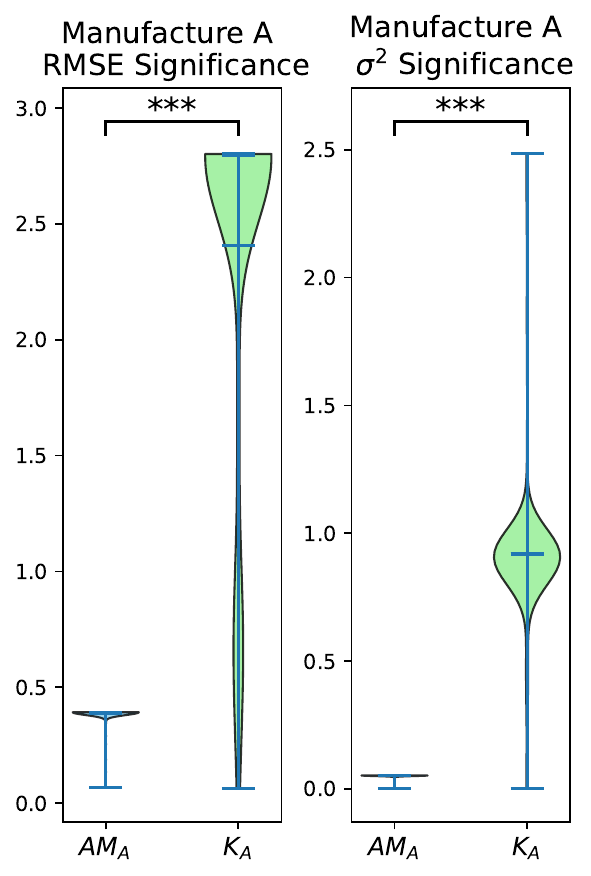}
    \put(-5,98){\footnotesize\textbf{II}}
\end{overpic}
\end{minipage}

\caption{Geometric A: I:Comparison RMSE and $\sigma^2$ of Actual, Ideal Kneading, AM and Target PCL; II: RMSE and $\sigma^2$ Significance Analysis}
\label{Geometric A actual-ideal-scale}
\end{figure}

The quantitative ICP registration results for Geometry A are shown in Fig.~\ref{Geometric A actual-ideal-scale}. Before compensation, the actual kneaded PCL requires a relatively large Threshold to reach a fitness of 1.0, and the corresponding $\sigma^2$ exhibits irregular fluctuations as the Threshold increases. After compensation, the RMSE variation curve becomes smoother and approaches that of the ideal kneading result, while $\sigma^2$ converges more steadily. The significance analysis further confirms that the differences between the 3D-printed reference PCL and the actual kneading PCL are statistically significant in both RMSE and $\sigma^2$.

\begin{figure}[H]
\centering
\includegraphics[width=0.75\columnwidth]{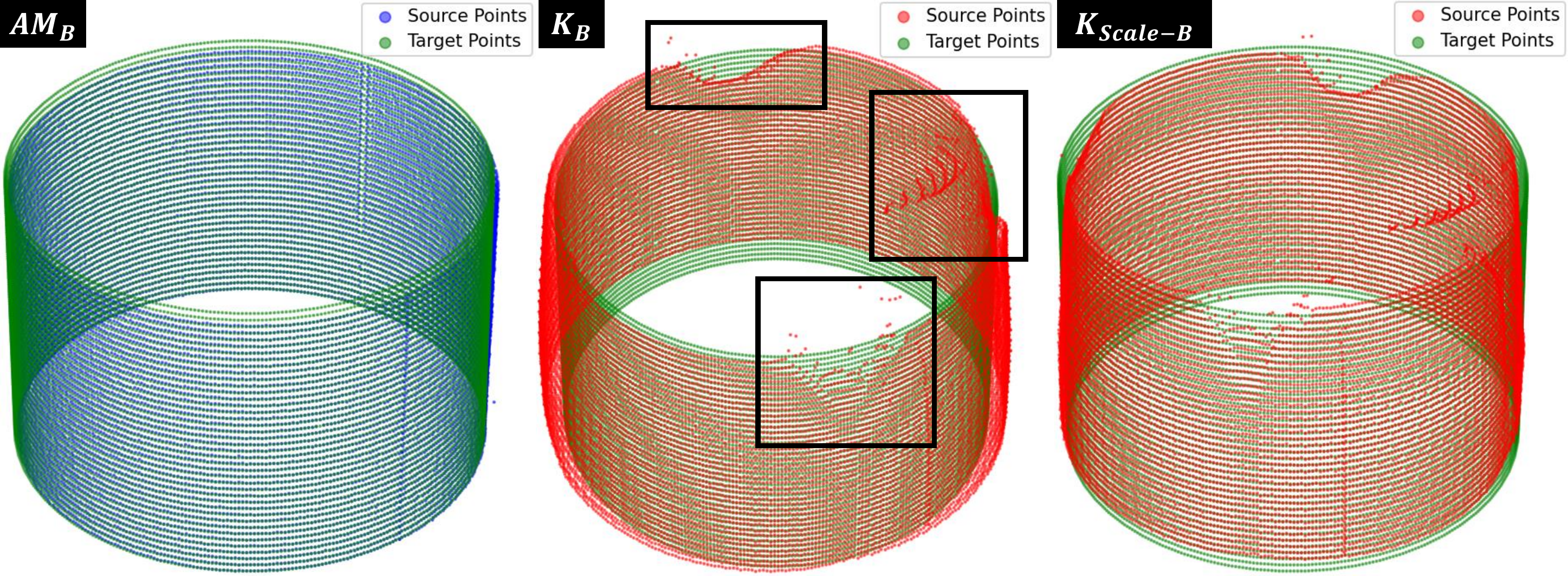}
\caption{\small Geometric B: Registration results between Actual knead, Compensate PCL and Target PCL}
\label{Geometric B actual and scale PCL}
\end{figure}

In contrast, Geometry B demonstrates a different error pattern. As illustrated in Fig.~\ref{Geometric B actual and scale PCL}, although the overall cylindrical shape is successfully formed, local corner-related defects are observed in the actual kneaded PCL. These defects, highlighted by the black boxes, originate from the initial square blank, where insufficient material redistribution at the corner regions leads to partial contour loss during the transition to a circular geometry. Such local defects persist even after compensation, although the global geometric consistency of the PCL is improved.

\begin{figure}[htbp]
\centering

\begin{minipage}{0.75\textwidth}
\centering
\begin{overpic}[width=\linewidth]{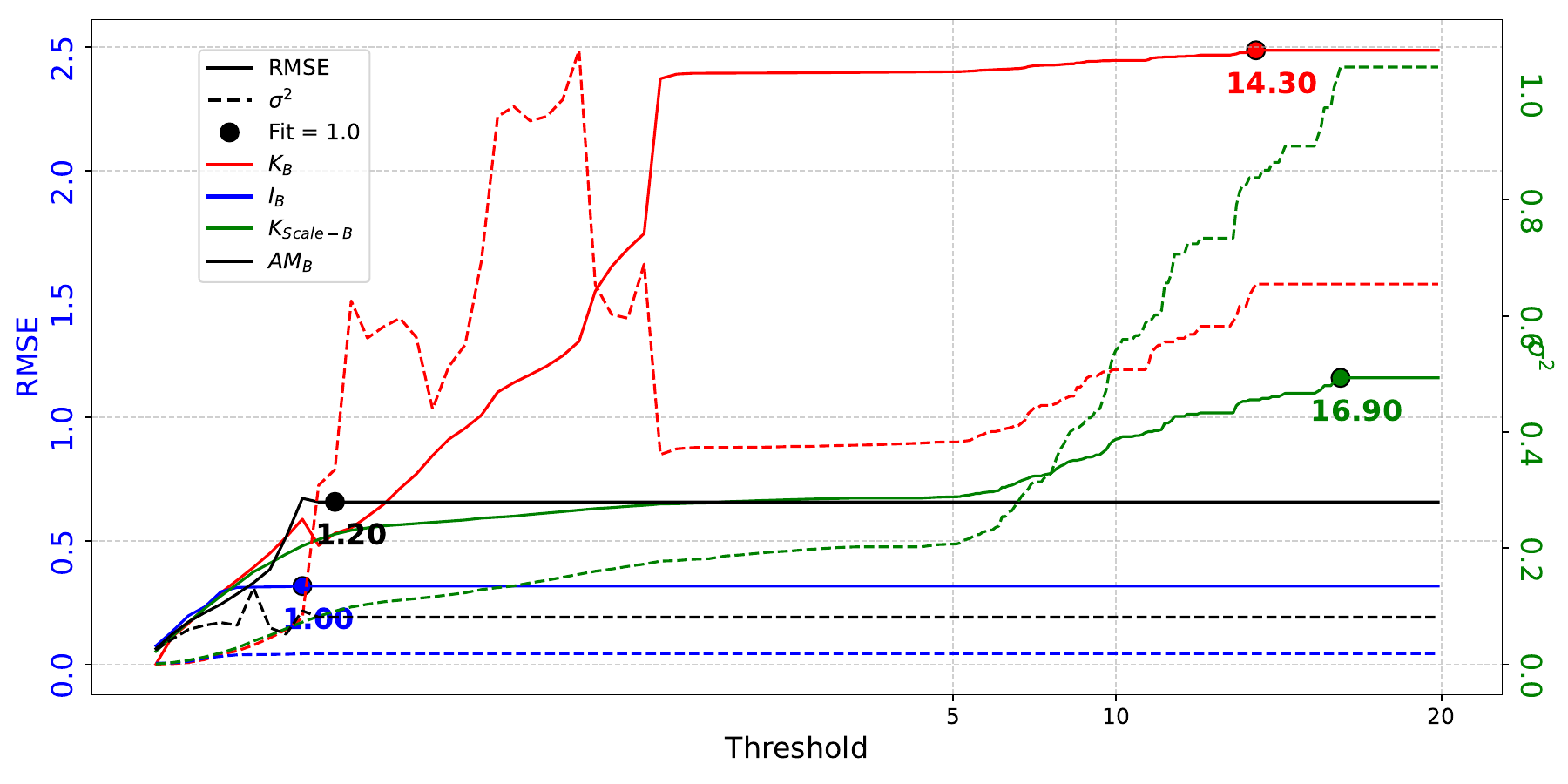}
    \put(0,50){\footnotesize\textbf{I}}
\end{overpic}
\end{minipage}\hfill
\begin{minipage}{0.25\textwidth}
\centering
\begin{overpic}[width=\linewidth]{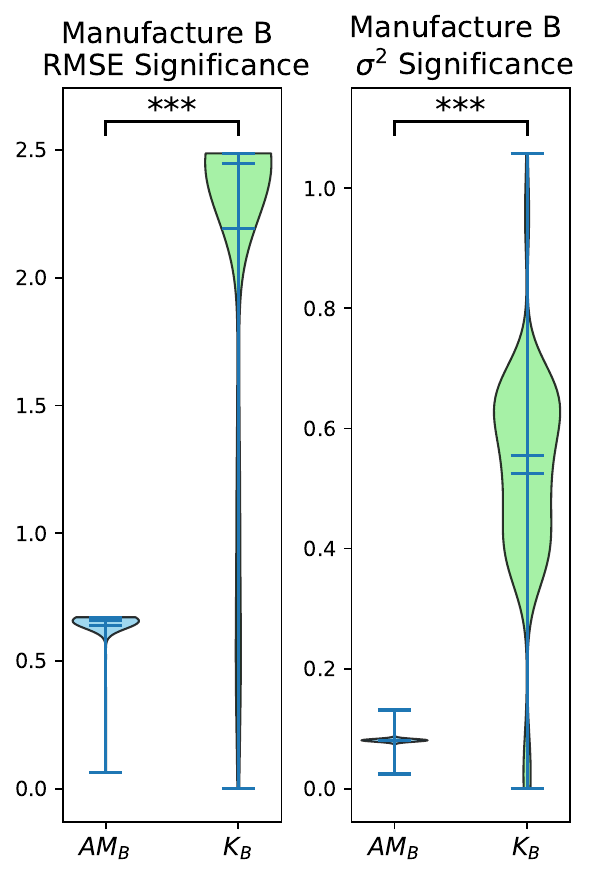}
    \put(-5,98){\footnotesize\textbf{II}}
\end{overpic}
\end{minipage}

\caption{Geometric B: I:Comparison RMSE and $\sigma^2$ of Actual, Ideal Kneading, AM and Target PCL; II: RMSE and $\sigma^2$ Significance Analysis}
\label{Geometric B actual-ideal-scale}
\end{figure}

The corresponding ICP results in Fig.~\ref{Geometric B actual-ideal-scale} show that Geometry B reaches a fitness of 1.0 only at relatively large Threshold values before compensation. This behavior is mainly attributed to the incomplete PCL caused by corner defects, which hinders ICP convergence at smaller Thresholds. After compensation, the RMSE is reduced and the RMSE curve becomes smoother; however, the residual RMSE remains larger than that of Geometry A. Notably, Geometry B exhibits a smaller $\sigma^2$ than Geometry A, indicating that apart from the localized corner defects, the remaining PCL is highly consistent with the target geometry. The significance analysis further verifies that the RMSE and $\sigma^2$ differences between the kneaded PCLs and the reference PCL are statistically significant.

\begin{figure}[H]
\centering
\includegraphics[width=1.0\columnwidth]{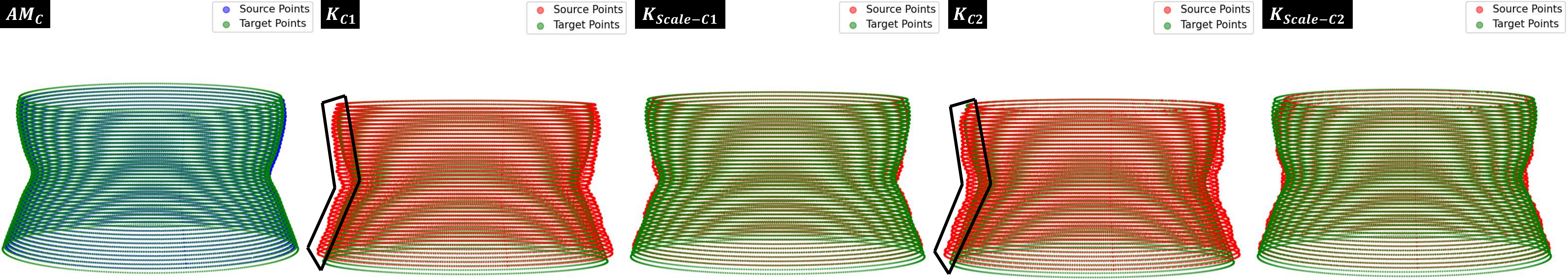}
\caption{\small Geometric C: Registration results between Actual knead, Compensate PCL and Target PCL}
\label{Geometric C actual and scale PCL}
\end{figure}

Geometries C, D, and E represent geometries with continuous or quasi-continuous axial characteristics and were formed using two kneading strategies: the Similar Gradient method and the Envelope Shaping First method. The corresponding actual kneaded PCLs, compensated PCLs, and target PCLs are shown in Fig.~\ref{Geometric C actual and scale PCL}, Fig.~\ref{Geometric D actual and scale PCL}, and Fig.~\ref{Geometric E actual and scale PCL}, respectively.

For Geometry C, the deviation between the actual kneaded PCL and the target PCL exhibits a gradual axial variation rather than a uniform scale expansion. As shown in Fig.~\ref{Geometric C actual and scale PCL}, the Similar Gradient method yields a PCL that more closely follows the axial geometric evolution of the target shape, whereas the Envelope Shaping First method shows larger local deviations. After RMSE-based compensation, both methods achieve improved geometric alignment; however, the compensated PCL obtained using the Similar Gradient method exhibits lower RMSE and faster convergence of $\sigma^2$.

\begin{figure}[htbp]
\centering

\begin{minipage}{0.75\textwidth}
\centering
\begin{overpic}[width=\linewidth]{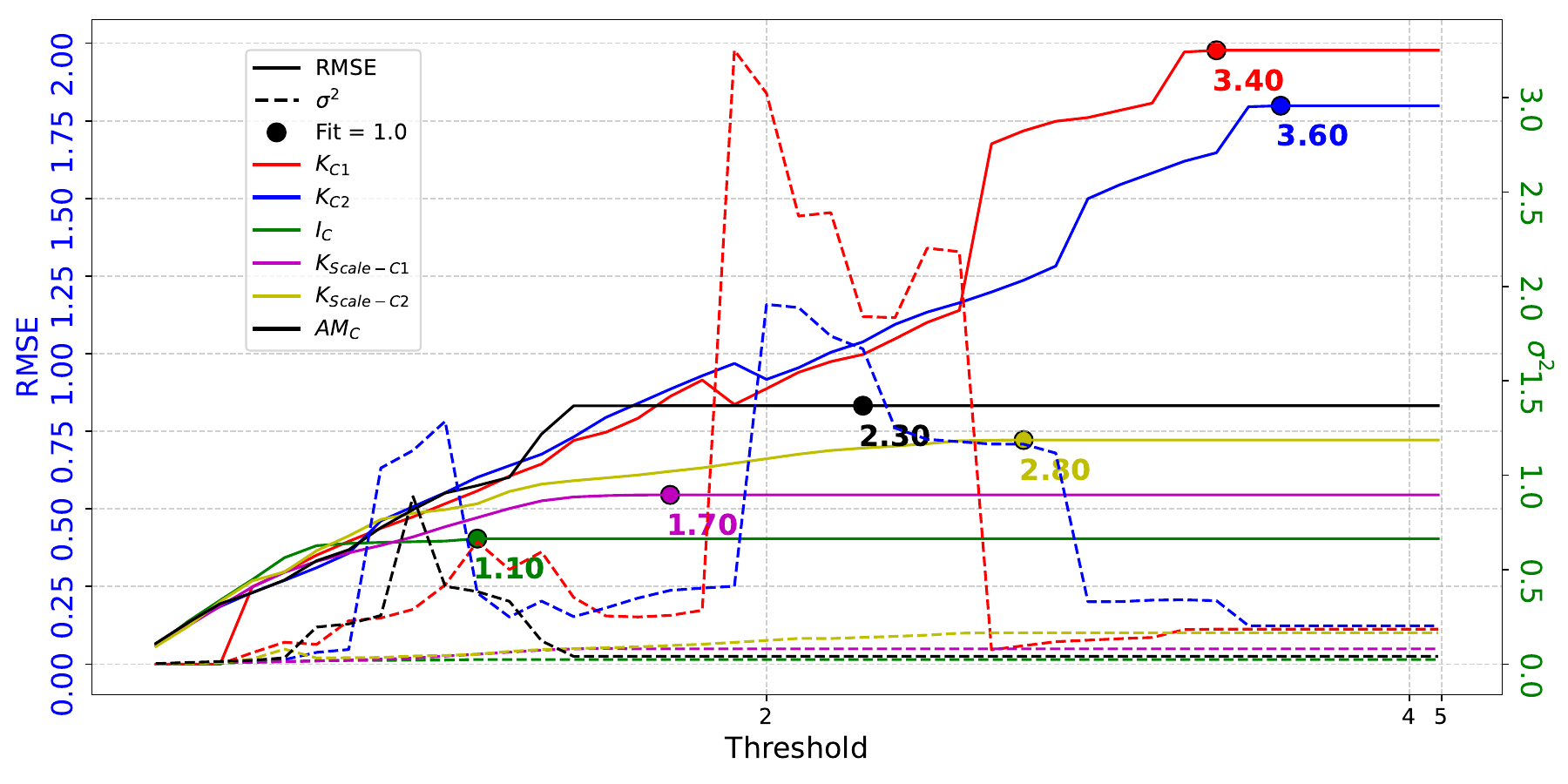}
    \put(0,50){\footnotesize\textbf{I}}
\end{overpic}
\end{minipage}\hfill
\begin{minipage}{0.25\textwidth}
\centering
\begin{overpic}[width=\linewidth]{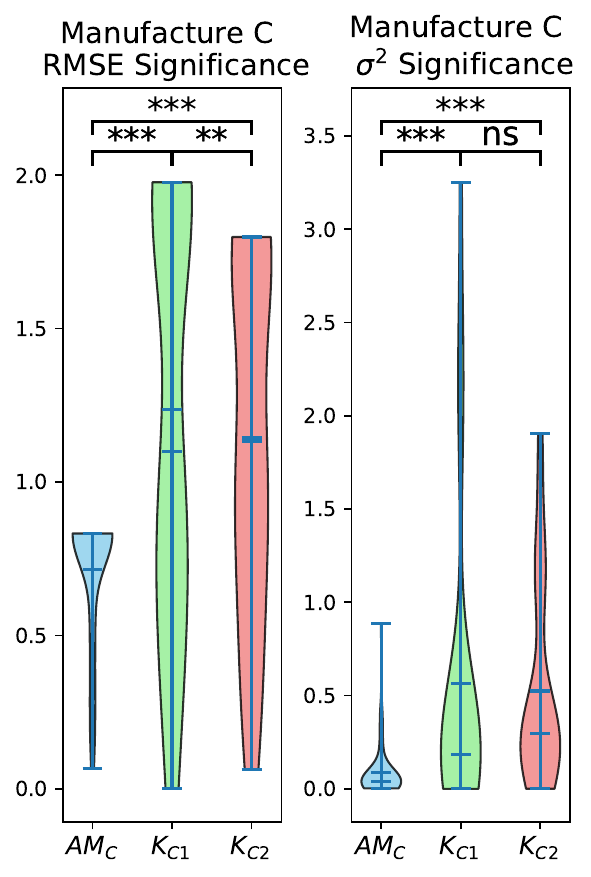}
    \put(-5,98){\footnotesize\textbf{II}}
\end{overpic}
\end{minipage}

\caption{Geometric C: I:Comparison RMSE and $\sigma^2$ of Actual, Ideal Kneading, AM and Target PCL; II: RMSE and $\sigma^2$ Significance Analysis}
\label{Geometric C actual-ideal-scale}
\end{figure}

\begin{figure}[H]
\centering
\includegraphics[width=1.0\columnwidth]{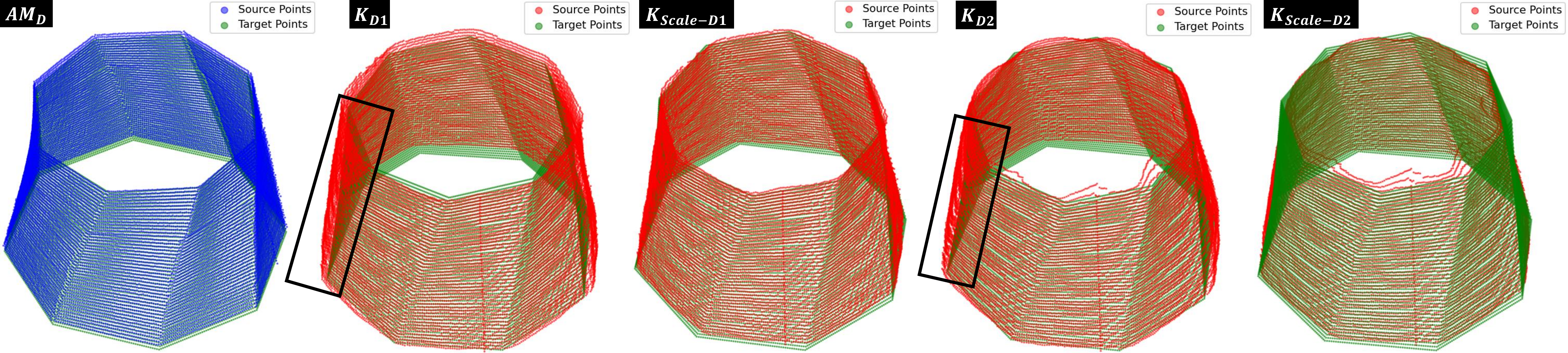}
\caption{\small Geometric D: Registration results between Actual knead, Compensate PCL and Target PCL}
\label{Geometric D actual and scale PCL}
\end{figure}

This trend is quantitatively confirmed by the ICP results in Fig.~\ref{Geometric C actual-ideal-scale}. Before compensation, the Similar Gradient method reaches a fitness of 1.0 at a smaller Threshold than the Envelope Shaping First method. After compensation, this difference becomes more pronounced, with the compensated Similar Gradient PCL approaching the ideal kneading accuracy. The corresponding significance analysis indicates that the differences in RMSE between the two methods are statistically significant, while the differences in $\sigma^2$ are less pronounced, suggesting that the primary distinction lies in systematic geometric deviation rather than random dispersion.

Geometry D exhibits behavior dominated by global elastic rebound, similar to Geometry A, but without discrete cross-sectional transformation. As shown in Fig.~\ref{Geometric D actual and scale PCL}, both kneading methods result in an overall scale expansion of the PCL relative to the target PCL. The black-box regions indicate that the deviation is spatially uniform rather than localized. After compensation, both methods show a substantial improvement in overlap between the compensated PCLs and the target PCL.

\begin{figure}[htbp]
\centering

\begin{minipage}{0.75\textwidth}
\centering
\begin{overpic}[width=\linewidth]{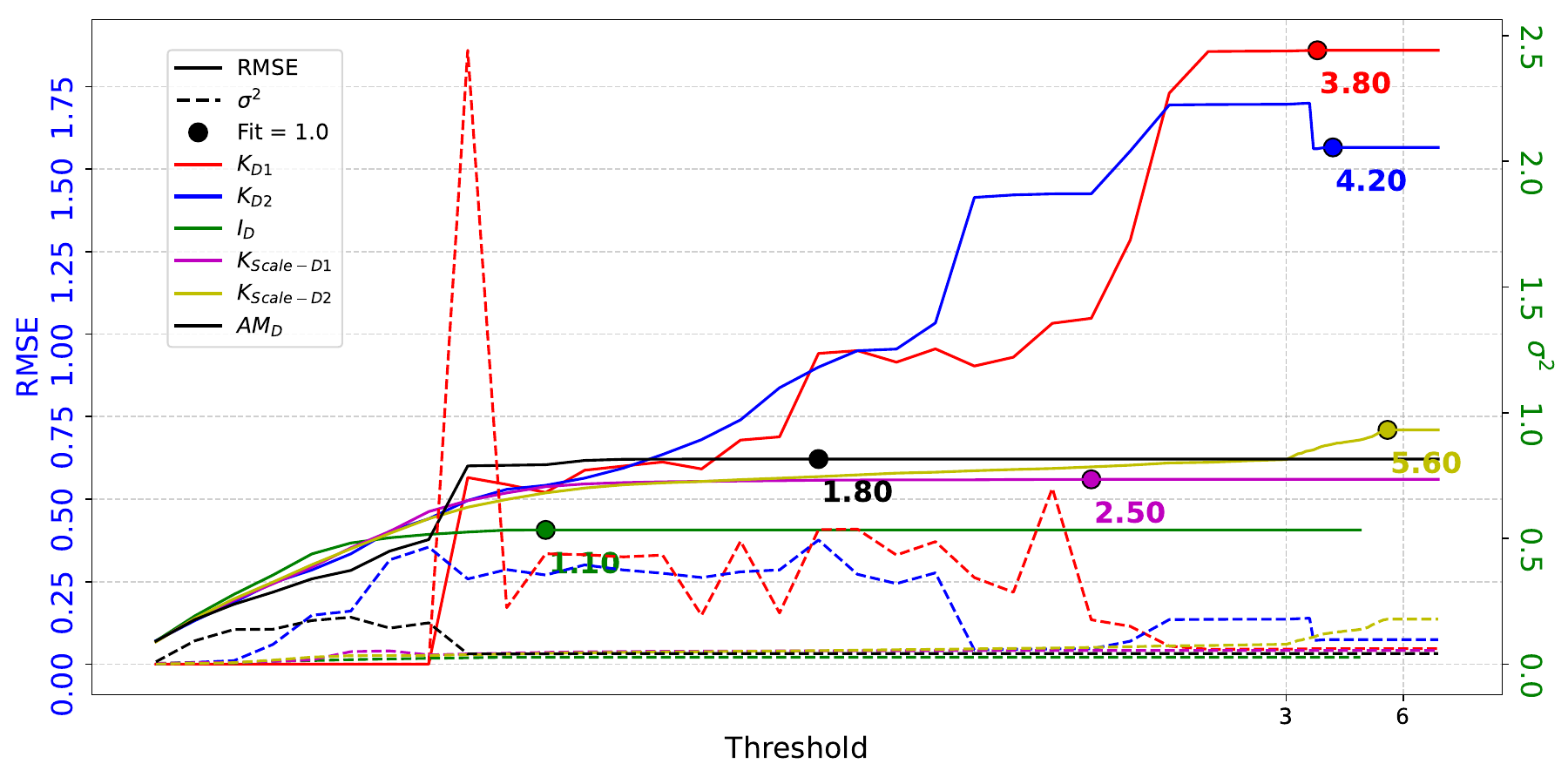}
    \put(0,50){\footnotesize\textbf{I}}
\end{overpic}
\end{minipage}\hfill
\begin{minipage}{0.25\textwidth}
\centering
\begin{overpic}[width=\linewidth]{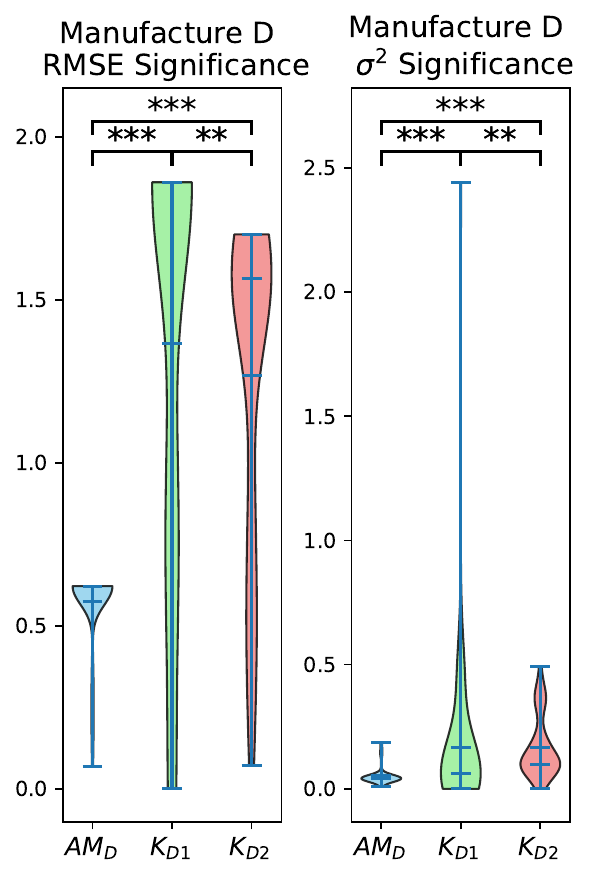}
    \put(-5,98){\footnotesize\textbf{II}}
\end{overpic}
\end{minipage}

\caption{Geometric D: I:Comparison RMSE and $\sigma^2$ of Actual, Ideal Kneading, AM and Target PCL; II: RMSE and $\sigma^2$ Significance Analysis}
\label{Geometric D actual-ideal-scale}
\end{figure}

\begin{figure}[H]
\centering
\includegraphics[width=1.0\columnwidth]{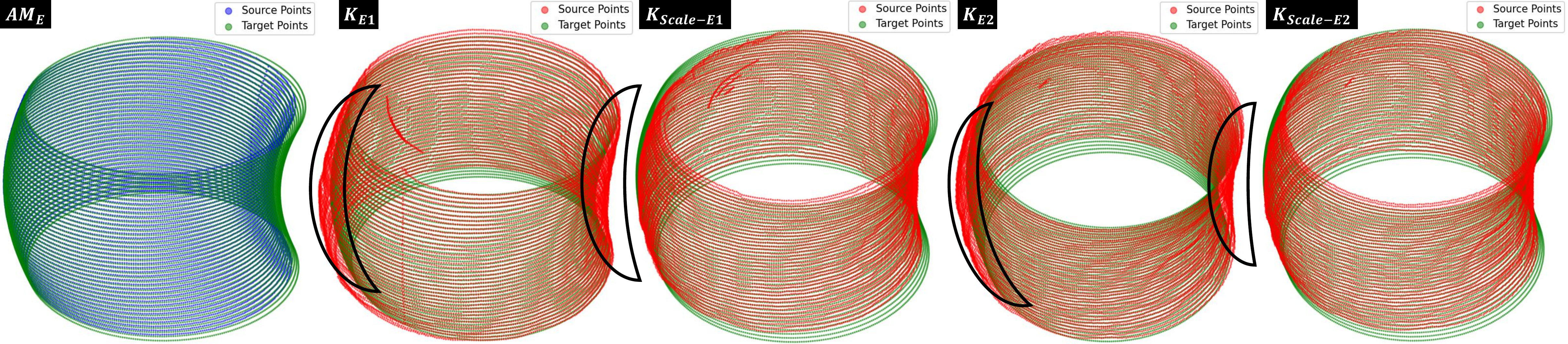}
\caption{\small Geometric E: Registration results between Actual knead, Compensate PCL and Target PCL}
\label{Geometric E actual and scale PCL}
\end{figure}

The ICP results in Fig.~\ref{Geometric D actual-ideal-scale} reveal that compensation leads to smoother RMSE variation curves and reduced $\sigma^2$ values for both methods. Compared with Geometry C, Geometry D requires larger Threshold values to reach a fitness of 1.0, reflecting its larger overall geometric deviation. The significance analysis confirms that both kneading methods differ significantly from the 3D-printed reference PCL in terms of RMSE and $\sigma^2$, while the difference between the two methods is mainly reflected in the stability of the error distribution.

Geometry E exhibits a distinctly different error mechanism compared with Geometries C and D. As shown in Fig.~\ref{Geometric E actual and scale PCL}, the primary deviation is not global scale expansion but axial center offset of identical cross-sections in the PCL. Despite having identical cross-sectional shapes and dimensions along the axis, the centers of these cross-sections deviate systematically, leading to local misalignment that cannot be eliminated through global compensation.

\begin{figure}[htbp]
\centering

\begin{minipage}{0.75\textwidth}
\centering
\begin{overpic}[width=\linewidth]{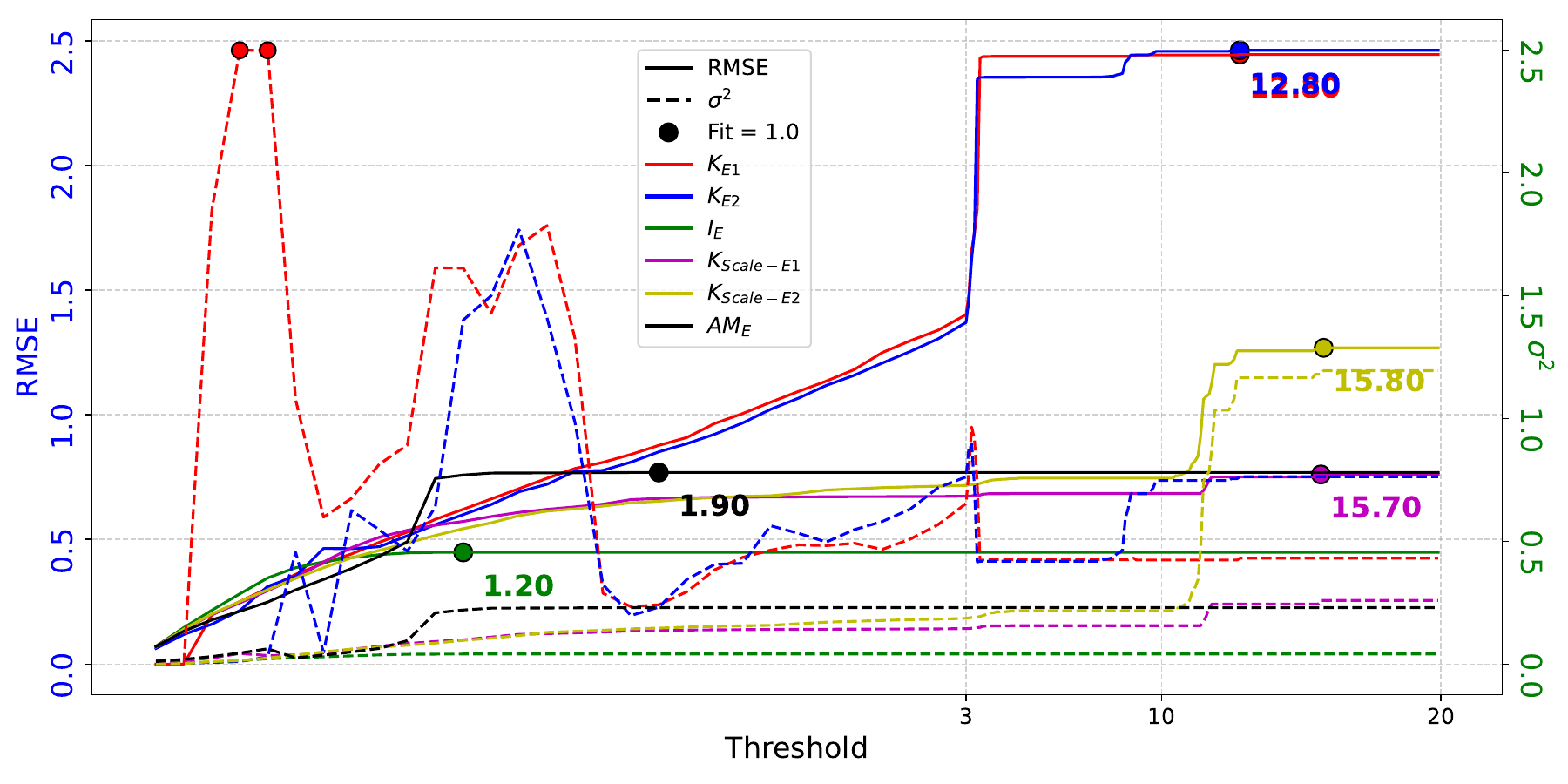}
    \put(0,50){\footnotesize\textbf{I}}
\end{overpic}
\end{minipage}\hfill
\begin{minipage}{0.25\textwidth}
\centering
\begin{overpic}[width=\linewidth]{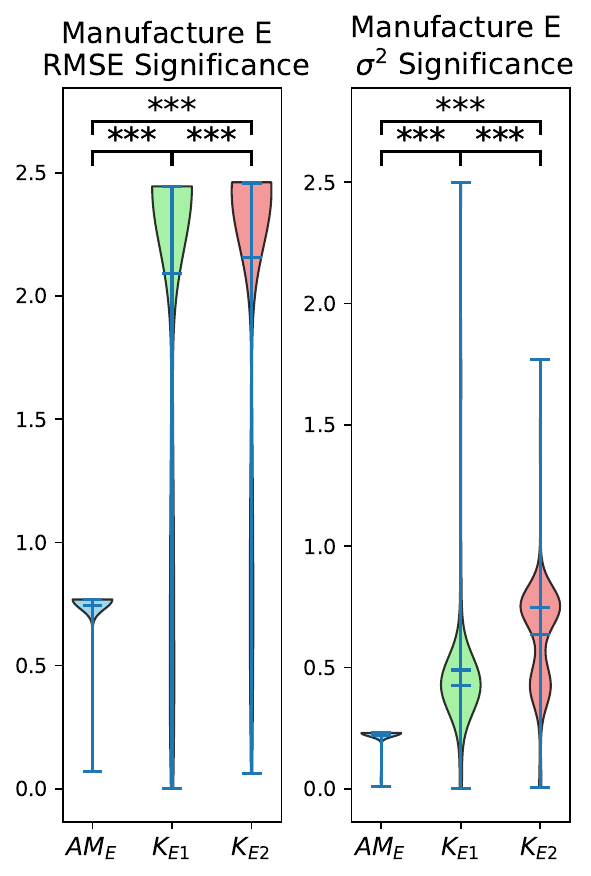}
    \put(-5,98){\footnotesize\textbf{II}}
\end{overpic}
\end{minipage}

\caption{Geometric E: I:Comparison RMSE and $\sigma^2$ of Actual, Ideal Kneading, AM and Target PCL; II: RMSE and $\sigma^2$ Significance Analysis}
\label{Geometric E actual-ideal-scale}
\end{figure}

This behavior is further illustrated in Fig.~\ref{Geometric E actual-ideal-scale}. Before compensation, the RMSE curves of both kneading methods remain relatively stable, while $\sigma^2$ exhibits large fluctuations at small Threshold values. After compensation, the Threshold values corresponding to a fitness of 1.0 remain nearly unchanged, indicating that RMSE-based global compensation has limited effectiveness for Geometry E. The significance analysis confirms that both kneading methods differ significantly from the reference PCL, and that the dominant error source is related to kneading command planning rather than material rebound.

\begin{table}[H]
\centering
\caption{Geometry ICP Registration between Actual Kneading and Compensated PCL Data}
\begin{tabular}{ccccc}
\hline
Geometric          & Kneading Method & Threshold & RMSE  & $\sigma^2$    \\
\multirow{3}{*}{A} & $K_A$              & 14.8      & 2.802 & 0.937 \\
                   & $I_A$              & 2.3       & 0.895 & 0.234 \\
                   & $K_{Scale-A}$        & 16.9      & 1.202 & 0.843 \\
\hline
\multirow{3}{*}{B} & $K_B$              & 14.3      & 2.487 & 0.655 \\
                   & $I_B$              & 1         & 0.316 & 0.018 \\
                   & $K_{Scale-B}$        & 16.9      & 1.16  & 1.029 \\
\hline
\multirow{5}{*}{C} & $K_{C1}$              & 3.4       & 1.977 & 0.185 \\
                   & $K_{C2}$             & 3.6       & 1.799 & 0.203 \\
                   & $I_C$              & 1.1       & 0.404 & 0.024 \\
                   & $K_{Scale-C1}$       & 1.7       & 0.545 & 0.081 \\
                   & $K_{Scale-C2}$       & 2.8       & 0.722 & 0.165 \\
\hline
\multirow{5}{*}{D} & $K_{D1}$             & 3.8       & 1.86  & 0.061 \\
                   & $K_{D2}$             & 4.2       & 1.567 & 0.097 \\
                   & $I_D$              & 1.1       & 0.407 & 0.027 \\
                   & $K_{Scale-D1}$       & 2.5       & 0.56  & 0.054 \\
                   & $K_{Scale-D2}$       & 5.6       & 0.71  & 0.179 \\
\hline
\multirow{5}{*}{E} & $K_{E1}$             & 12.8      & 2.445 & 0.431 \\
                   & $K_{E2}$             & 12.8      & 2.462 & 0.763 \\
                   & $I_E$              & 1.2       & 0.448 & 0.041 \\
                   & $K_{Scale-E1}$       & 15.8      & 0.76  & 0.259 \\
                   & $K_{Scale-E2}$       & 15.7      & 1.268 & 1.195 \\
\hline
\end{tabular}
\label{Geometry ICP Registration between Actual Kneading and Compensated PCL Data}
\end{table}

The overall kneading accuracy of Geometries A--E is quantitatively summarized in Tab.~\ref{Geometry ICP Registration between Actual Kneading and Compensated PCL Data}. The results reveal that the dominant forming error and the effectiveness of RMSE-based compensation are strongly dependent on the geometric characteristics of the target shape. Geometry A, involving a discrete cross-sectional transformation from circular to square, exhibits the largest RMSE and unstable $\sigma^2$ due to amplified elastic rebound and corner-induced stress redistribution, while compensation effectively suppresses this global deviation. Geometry B, formed from a square blank into a circular geometry, is primarily limited by local corner-related material deficiency rather than global scale expansion, leading to reduced RMSE after compensation but persistent residual error. For Geometry C, characterized by continuous axial geometric variation, compensation significantly improves accuracy, with the Similar Gradient method achieving lower RMSE and faster $\sigma^2$ convergence than the Envelope Shaping First method. Geometry D is dominated by uniform elastic rebound, and compensation consistently reduces RMSE for both kneading strategies. In contrast, Geometry E shows limited sensitivity to global compensation, as its dominant error originates from axial center offset rather than material rebound. Overall, RMSE at fitness 1.0 serves as an effective indicator of rebound-dominated error, while the compensated RMSE represents the upper bound of achievable kneading accuracy under given material and environmental conditions.

\section{Conclusion}

This study proposes a volume-consistent kneading-based forming paradigm for plastic materials, integrating a custom-designed kneading platform, geometry-aware kneading command generation, and PCL-based geometric evaluation. By combining envelope-type and non-envelope-type kneading strategies with layer-wise scanning and ICP-based registration, the proposed system enables continuous shaping of complex three-dimensional geometries with high material utilization.

Experimental validation was conducted on five representative geometries (A–E) covering discrete cross-sectional transformations, continuous axial variations, and center-offset configurations. The results demonstrate that elastic rebound is the dominant source of global geometric deviation for most geometries, and that RMSE at a fitness of 1.0 provides an effective quantitative indicator of rebound-dominated error. RMSE-based compensation significantly improves the geometric fidelity of rebound-dominated geometries, and the compensated RMSE can be regarded as the upper bound of achievable kneading accuracy under given material and environmental conditions.

The comparative analysis further reveals clear geometry-dependent error mechanisms. Geometries involving discrete cross-sectional transformations (A and B) are strongly influenced by corner-induced stress redistribution and material deficiency, while geometries with continuous axial variation (C and D) benefit more consistently from compensation, particularly when combined with the Similar Gradient method. In contrast, Geometry E exhibits limited sensitivity to global compensation due to axial center offset, indicating that its dominant error originates from kneading command planning rather than material rebound.

Overall, the proposed kneading-based forming framework demonstrates high material efficiency (above 98\%), controllable geometric accuracy, and strong adaptability to different geometric characteristics. These results highlight the potential of kneading-based forming as a sustainable alternative to conventional additive and subtractive manufacturing for customizable and low-waste fabrication. Future work will focus on adaptive path re-planning and local compensation strategies to address center-offset and non-rebound-dominated geometries.

\appendix
\section{Example Appendix Section}
\label{app1}

Appendix text.

\bibliographystyle{elsarticle-num} 
\bibliography{references}

\end{document}